\documentclass[preprint,12pt]{elsarticle}

\usepackage{caption, subcaption}
\usepackage{graphicx}
\usepackage{xcolor}
\usepackage{url}
\usepackage{amssymb}
\usepackage[section]{placeins}

\journal{Neuroimage}

\begin{document}

\begin{frontmatter}


\title{Ensemble learning with 3D convolutional neural networks for functional connectome-based prediction}


 \author[label1]{Meenakshi Khosla}
  \author[label2]{Keith Jamison}
  \author[label2,label3]{Amy Kuceyeski}
   \author[label1,label4]{Mert R. Sabuncu}
\address[label1]{School of Electrical and Computer Engineering, Cornell University}
 \address[label2]{Radiology, Weill Cornell Medical College}
\address[label3]{Brain and Mind Research Institute, Weill Cornell Medical College}
\address[label4]{Nancy E. \& Peter C. Meinig School of Biomedical Engineering, Cornell University}

\begin{abstract}
The specificty and sensitivity of resting state functional MRI (rs-fMRI) measurements depend on preprocessing choices, such as the parcellation scheme used to define regions of interest (ROIs). 
In this study, we critically evaluate the effect of brain parcellations on machine learning models applied to rs-fMRI data. 
Our experiments reveal an intriguing trend: On average, models with stochastic parcellations consistently perform as well as models with widely used atlases at the same spatial scale. 
We thus propose an ensemble learning strategy to combine the predictions from models trained on connectivity data extracted using different (e.g., stochastic) parcellations.
We further present an implementation of our ensemble learning strategy with a novel 3D Convolutional Neural Network (CNN) approach. 
The proposed CNN approach takes advantage of the full-resolution 3D spatial structure of rs-fMRI data and fits non-linear predictive models. 
Our ensemble CNN framework overcomes the limitations of traditional machine learning models for connectomes that often rely on region-based summary statistics and/or linear models. 
We showcase our approach on a classification (autism patients versus healthy controls) and a regression problem (prediction of subject's age), and report promising results. 
\end{abstract}


\end{frontmatter}

%\linenumbers

\section{Introduction}
\label{S:1}
%\subsection{Background}
Functional connectivity, as often captured by correlations in resting state functional MRI (rs-fMRI) data, has produced novel insights linking differences in brain organization to individual or group-level characteristics. 
Recently, machine learning models are being increasingly applied to study and exploit individual variation in functional connectivity data~\cite{Plitt2015,Mennes2012,Varoquaux2010}. 
These models often employ hand-engineered features, such as pairwise correlations between regions of interest (ROIs) and network topological measures of clustering, modularity, small-worldness, integration, or segregation~\cite{BrownH16,Kaiser2011, AlexanderBloch2013}. 
%Practically all of these methods rely on prior definitions of network nodes, defined using a brain parcellation scheme or atlas. 
The ROIs are usually computed based on a pre-defined atlas or a parcellation scheme.  
The choice of the ROIs can have a significant impact on downstream analyses~\cite{Smith2011,Yao2015,dadi2018}. 
%The choice of the ``optimal'' atlas for the classification task is far from straightforward and is often made arbitrarily in the rs-fMRI community. 
%Variability in precise node definitions across individuals further complicates the task of choosing a brain atlas that is well-representative of the population. 
%However, significant differences have been found in the analysis of brain networks depending on the choice of brain atlas \cite{Yao2015,dadi2018}. 
%[WRITE about the methods of generating atlases: anatomical and functional and the pros/cons of each]

Brain ROIs can be defined based on macro-anatomical features, cytoarchitecture, functional activations, and/or connectivity patterns~\cite{fischl2002whole,glasser2016multi,eickhoff2015connectivity,ARSLAN20185}. 
%The latter approach is more recent, and is believed to yield coherent functional units that represent the underlying functional connectome more faithfully~\cite{ARSLAN20185}. 
A common approach is to derive the ROIs either based on input from experts and/or using a data-driven strategy on a small number of subjects.
Expert-defined ROIs are challenging to standardize across studies~\cite{yushkevich2015quantitative} and often rely on arbitrary decisions.
Data-driven ROIs, on the other hand, can be biased by the selection of the subjects, especially for regions that 
exhibit large variability across the population. 
Popular data-driven techniques include clustering, dictionary learning and Independent Component Analysis (ICA)~\cite{varoquaux2011multi,Yeo2011,dadi2018}. 
Such methods can be sensitive to confounds such as motion, while initialization, optimization, and other algorithmic choices can also significantly influence the results~\cite{thirion2014fmri}.
A parcellation scheme not only defines the boundaries of ROIs, but also restricts the analysis to a certain spatial scale. Abraham et al. \cite{ABRAHAM2017} showed that among various preprocessing decisions, the choice of region definition has the greatest impact on predictive accuracy with data-driven extraction based on dictionary learning outperforming ICA/clustering and other reference atlases.
%However, the scale at which pathological effects can be discerned is rarely known \textit{a priori}. 
%and the network resolution or atlas is usually chosen in an ad-hoc manner. 
%Thus, we anticipate that connectome-based prediction models can benefit markedly from an ensemble strategy that integrates across different scales and ROI definitions. 
%Figure \ref{fig:pipeline} shows a general schematic of our proposed framework. 
%Our approach allows us to probe functional brain networks at multiple scales and find the optimal combination for classification. 
%Through a rigorous empirical analysis, we demonstrate that model averaging over stochastic parcellations consistently yields better performance than single atlas-based approaches. 
%Furthermore, our stochastic multi-parcellation approach is ``scale-agnostic'' and the ensemble reliably performs as well as the best classification scale. 

%The idea of ensemble learning over different parcellation schemes is motivated by a strong theoretical basis (add equations?) It helps in overcoming the implicit bias introduced by choosing to represent the data in a single ad-hoc way. (one representation of the data over others)

\begin{figure}[t]
\hspace{-0.3in}
\centering
\includegraphics[width=1.1\linewidth]{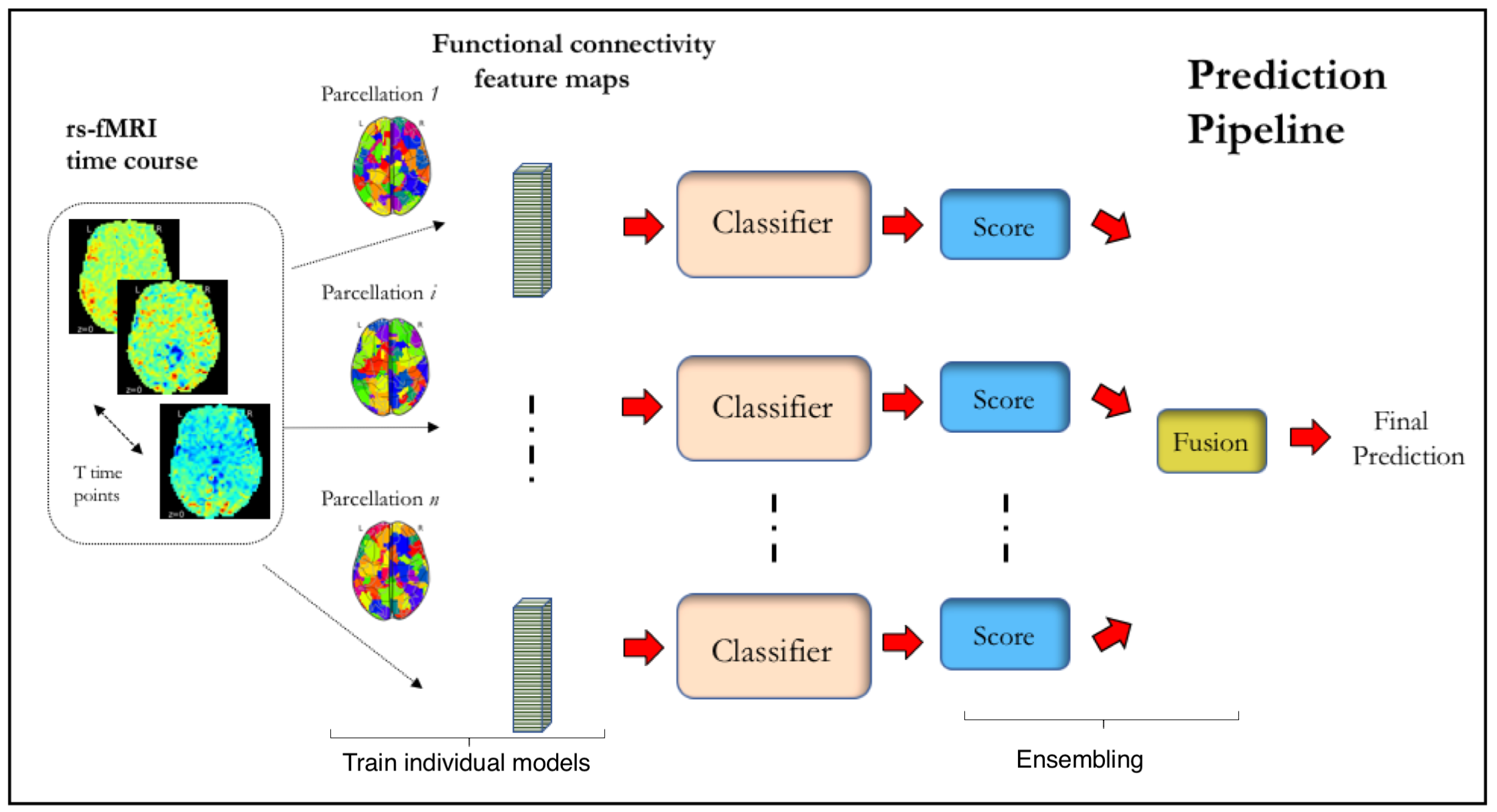}
\caption{A general illustration of the proposed approach}
\label{fig:pipeline}
\end{figure}

Given the arbitrary nature of a chosen parcellation scheme and its impact on predictive models, we hypothesized that machine learning models can benefit markedly from an ensemble strategy that integrates across different scales and ROI definitions.
Figure~\ref{fig:pipeline} shows a general schematic of our proposed framework. 
%% Comment on surface-based versus volume-based parcellations. 
%Several studies have highlighted the critical role of network node definition on connectivity analysis~\cite{marrelec2011assessing}.
%The impact of the choice of ROI-definitions on predictive models, particularly in the context of deep learning, is poorly understood. 
%Most large-scale benchmark studies thus far have been limited in focus to more traditional machine learning models for classification. 
In this work, we conducted a thorough empirical evaluation  of different choices for brain parcellations.
%and classification algorithms. 

Another important factor in connectome-based machine learning pertains to the choice of the classification algorithm. 
A large body of related work in the literature has focused on simple linear predictive models using vectorized connectivity data. A relatively recent trend is to exploit neural networks for graph-structured data, such as Graph Convolution Networks or BrainNet-CNN, to make individual-level predictions on connectomes. Ktena et al. \cite{KTENA2018} applied spectral graph convolutions in a distance-metric learning framework to train a k-nearest neighbor classifier on connectivity data. In a similar vein, Kawahara et al. \cite{Kawahara2017} proposed the BrainNetCNN architecture that extends convolutional neural networks (CNNs) to handle graph-structured data. 
CNNs are motivated via the translation-invariance property of image-based classification problems and can exploit voxel/pixel resolution data. On the other hand, BrainNetCNN works directly with an adjacency matrix derived from the connectome data, while disregarding spatial information. The model parameter count would scale according to the number of ROIs, making the utilization of voxel-level connectivity infeasible with this approach. As we discuss below, we propose an alternative representation of connectivity data, which allows us to leverage modern deep learning architectures, like CNNs, to build a prediction model that exploits the full-resolution 3D spatial structure of rs-fMRI without having to learn too many model parameters. 
%Henceforth, we critically evaluate our proposed 3D CNN framework for connectome-based classification. 

In this work, we consider two applications: discrimination of autism patients and healthy controls; and regression of age.
The first problem is a particularly challenging one.
Several previous studies have reported altered functional connectivity patterns in Autism Spectrum Disorder (ASD) patients~\cite{Cherkassky2006FunctionalCI,assaf2010,MONK2009764,Heinsfeld18}. While studies using small samples have reported classification accuracies over 75\% \cite{Yahata2016}, application of similar models on large heterogeneous datasets, such as ABIDE~\cite{DiMartino2017}, have shown more modest performance levels over a wide range of connectome preprocessing schemes (accuracies that range 60-67\%) \cite{ABRAHAM2017}. 
%We conjecture that these inadequacies can be addressed by innovative machine learning techniques for connectomes. 
%Diagnostics based on these neuroimaging biomarkers could lead to early disease identification, well before behavioral symptoms emerge. This is consequential, since early intervention in various psychiatric disorders holds significant therapeutic potential. 

%\subsection{Contribution}

Our main contributions in this paper are:
\begin{itemize} 
\item An extensive evaluation of the influence of brain parcellations on functional connectome-based machine learning models
\item An ensemble learning strategy for combining predictions from multiple classifiers corresponding to different brain parcellations 
\item An easy-to-implement 3D CNN framework for connectome-based classification 
\end{itemize}

\section{Materials and Methods}
\label{S:2}

\subsection{Dataset}
The Autism Brain Imaging Data Exchange (ABIDE) is a multi-site consortium aggregating and openly sharing anatomical, functional MRI and phenotypic datasets of individuals diagnosed with ASD, as well as healthy controls (HC)~\cite{DiMartino2017}. The first phase of ABIDE (ABIDE-I) collected data from 1,112 individuals, comprising 539 individuals diagnosed with ASD and 573 typical controls across 17 sites. The second phase (ABIDE-II) aggregated 1,114 additional datasets, comprising 521 individuals with ASD and 593 healthy controls across 19 sites.

%\subsubsection{Cohort Selection} 

%For ABIDE-I, subject data that passed quality control assessment by all the functional raters was used in our study. This yielded a final sample size of 774 subjects, comprising 379 subjects with ASD and 395 typical controls. 

%\subsubsection{Independent test dataset}
%A cohort of 393 subjects from ABIDE-II, comprising 163 individuals with ASD and 230 typical controls, was used for independent testing. This included all subjects from sites that participated in ABIDE-I and used the same sequence parameters for data collection in ABIDE-II. 

\subsection{Preprocessing of fMRI Data}
The Preprocessed Connectomes Project (PCP) released preprocessed versions of ABIDE-I using several pipelines \cite{Cameron2013}. We used the data processed through the Configurable Pipeline for the Analysis of Connectomes (CPAC). This pipeline performs motion correction, global mean intensity normalization and standardization of functional data to MNI space (3x3x3 mm resolution) before the extraction of ROI time series. Among the different strategies in the release, our analysis used data de-noised by regression of nuisance signals including motion parameters, CompCor WM+CSF components, and global signal, followed by band-pass filtering (0.01-0.1Hz). We note that we have experimented  with alternate preprocessing strategies that include/exclude the global signal regression and CompCor steps. These results are presented in the Supplementary Section 7.6.

We preprocessed the ABIDE-II dataset following the same sequence of steps listed for ABIDE-I in CPAC (using the version v1.0.2a). 
Since manual quality control (QC) was not yet available for ABIDE-II, we performed an automatic QC by selecting those subjects that retained at least 100 frames or 4 minutes of fMRI scans after motion scrubbing~\cite{Power2014}. Motion scrubbing was performed based on Framewise Displacement (FD), discarding one volume before and two volumes after the frame with FD exceeding 0.5mm~\cite{Muschelli}. 

\subsection{Cohort selection}
In our experiments, we used ABIDE-I subject data that passed manual QC by all the functional raters. 
This yielded a final sample size of 774 ABIDE-I subjects, comprising 379 subjects with ASD and 395 typical controls. 
As an independent test dataset, we employed ABIDE-II subjects from sites that participated in ABIDE-I  and used the same MRI sequence parameters for data collection. After automatic QC, we ended up with a final ABIDE-II sample size of 163 individuals with ASD and 230 healthy controls. For age prediction, we only considered healthy controls. 
Furthermore, subjects whose age were more than 3.5 standard deviations away from the median were excluded from the task of age prediction. 
Table \ref{table:dataset} summarizes the dataset characteristics for the two prediction tasks considered in this study.

\begin{table}
 \begin{tabular}{||c c c c||} 
 \hline
Dataset & Prediction & Sample Size & Median Age (Range) in yrs\\ [0.5ex] 
 \hline\hline
 ABIDE-I & Age  &  387 & 13.8 (6.5-29.1) \\ 
 \hline
 ABIDE-I & ASD/HC  &  379/395  & 13.9 (6.5-56.2) \\
 \hline
 ABIDE-II & Age &  213 & 10.6 (5.8-18.8) \\
 \hline
 ABIDE-II & ASD/HC & 163/230  & 11.0 (5.2-38.9)  \\
 \hline
\end{tabular}
\caption{Composition of Cohorts}
\label{table:dataset}
\end{table}
%Patient demographics are summarized in Table \ref{table:}---Needed?

\subsection{Extracting ROI time series from atlases}
In our experiments, we considered all atlases that were used for ROI time series extraction in PCP. These include the following seven atlases: Talaraich and Tournoux (TT, R=97), Harvard-Oxford (HO, R=111), Automated Anatomical Labelling (AAL, R=116), Eickhoff-Zilles (EZ, R=116), Dosenbach 160 (DOS160, R=161), Craddock 200 (CC200, R=200), and Craddock 400 (CC400, R=392), where R is the number of ROIs~\cite{HO2, HO3,HO4, Dosenbach1358, DESIKAN2006968, AAL,CC,TT,EZ}.
%KJ: Worth mentioning: These are all discrete non-overlapping parcellations. ROI time series are simple voxel averages. Next part mentions connectivity matrices but no mention of how time series->connectivity was computed. Since this section is setting up the issues of ROI idiosyncracies, selection, scale etc.., perhaps include basics of how many ROIs each atlas contains, as well as the average and/or range of parcel volumes within each (some are fairly uniform, others vary widely). Others like the Dosenbach (I think?) have many ROIs but each is a small uniform sphere, with incomplete brain coverage. Should probably introduce the gray-matter masking here.
%\newline 

%KJ: I find this part confusing, especially before introducing the "connectivity fingerprint". Does it belong with the section on CNN? 
For our 3D CNN model, described below, the parcellated regions were used as target ROIs to derive the input connectivity features at the voxel level.
For the non-CNN benchmark models, also described below, each atlas was used to define a corresponding connectivity matrix which was fed as input to each model after collapsing into a vector.
We report results for ensemble learning strategies as well, where we combined the predictions of models corresponding to individual atlases. 

%KJ: I would probably merge this into the above subsection, and move the above paragraph about connectivity features to their respective CNN or non-CNN sections.
%KJ: more clarity on gray matter voxels: "as defined by a given mask" is unclear. We defined separately for each of the PCP atlases for those comparisons? Or did we end up with one MNI-defined gray matter mask?
\subsection{Creating stochastic parcellations}
\begin{figure}[!htb]
\centering 
\hspace{-0.5in}
\begin{minipage}{0.9\textwidth}  
\includegraphics[width=1.2\linewidth]{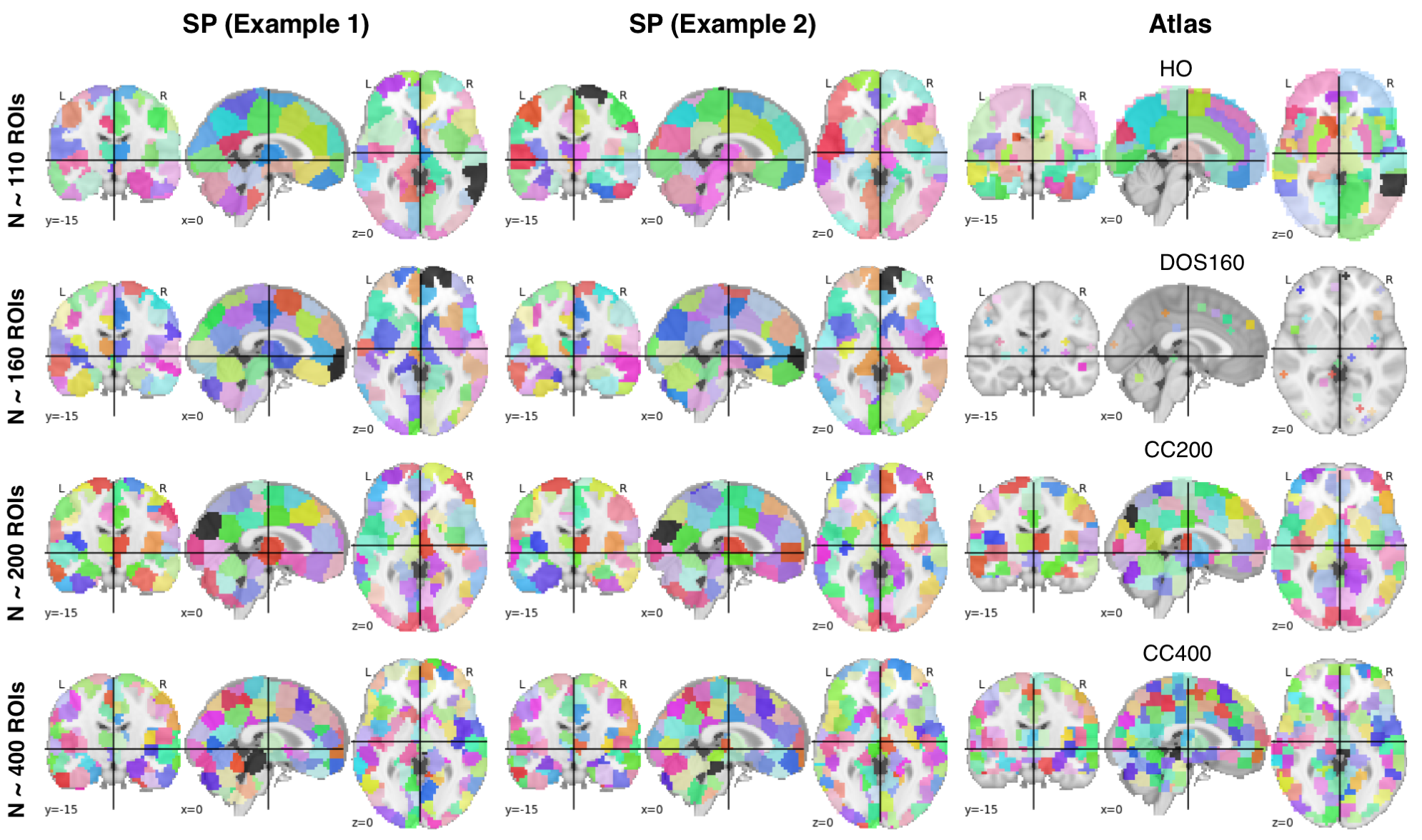}
%\caption{}
%\label{fig:cnn_violins_abide1_age}
\end{minipage}
\caption{ROI masks for example SPs and atlas at each of the four spatial scales considered in this study.}
\label{fig:masks}
\end{figure}

Stochastic parcellations were created by Poisson Disk Sampling using the method described in \cite{Schirmer}. Given a number of ROIs, this approach divides the gray matter voxels (as defined by a given mask) into roughly equal-sized parcels while ensuring that the parcels do not cross hemisphere boundaries. Stochasticity is introduced in the ROI center locations, and all the remaining voxels are assigned to the closest region center. These centers are kept a minimum distance apart based on the desired number of regions in the parcellation. Further details about the sampling approach are provided in Supplementary Section \ref{ssec:pds}. All parcellations were created in the MNI152 template at a 3mm resolution, same as the resolution of the preprocessed functional data. For creating these parcellations, we relied on a whole brain gray matter mask including sub-cortical structures. To create the mask, we took the union of the gray matter tissue prior provided in the standard MNI152 template and the cortical mantle mask used in \cite{Yeo2011}. Some example stochastic parcellations are shown in Figure \ref{fig:masks} against atlases at similar resolutions. 

%Further, we also created stochastic parcellations using the gray matter mask of individual atlases to ensure a fair comparison between the classification performance of atlases and stochastic parcellation approaches. 

%We created stochastic parcellations at 4 different scales, with the number of regions equal to 110, 160, 200 and 400 respectively. 
%For each scale, we created N=100 unique stochastic parcellations. 
%These scales were chosen in accordance with existing atlases.

%\subsection{From fMRI data to multi-channel connectivity fingerprints}

\subsection{3D Convolutional Neural Network Approach} \label{3dcnn}
Here, we present our novel strategy to adopt a 3D CNN architecture for use with connectomic data.

Loosely reminiscent of the biological visual system, CNNs use spatially localized filters to detect local image features. Unlike fully connected layers where every unit is connected to all other units of the previous layer, convolutional layers employ a structured arrangement where each unit is connected to only a small subset of spatially connected units in the input image channels. Further, the weights of these connections are shared between the units of the convolutional layer so that the same feature can be detected regardless of its spatial location. Mathematically, 
a convolutional layer of the form Y=$O_w$(X) operates on an M-dimensional input \textbf{X(v)=($X_1$(v),....,$X_M$(v))} by applying a set of filters \{W=\{$w_{m,n}$\}, \textit{m}=1,...,M; \textit{n}=1,...,N\}. Here, \textbf{v} is used to index the pixel or voxel (in case of 3D convolution). After applying an elementwise non-linearity $\phi$ (such as a logistic function) , this produces an N-dimensional output  \textbf{Y(v)=($Y_1$(v),....,$Y_N$(v))}. Each element $Y_n$(v), known as a feature map, is thus given as, \\
\begin{equation}
 Y_n(v)=\phi (\sum_{m=1}^{M}(X_m * w_{m,n})(v)),  
\end{equation}

where * denotes the standard spatial convolution operation.% and is given as $(X*w)(v) = \sum_{\tau} X(v-\tau ) w(\tau ) $. 
The convolutional layers in CNNs are often interspersed with pooling layers that reduce the size of feature maps and offer translation invariance. Max-pooling is the most popular pooling operation. It down-samples each input feature map (commonly referred to as a channel) separately by selecting the maximum feature response in pre-fixed local neighborhoods.  A max-pooling $Y_i=P(X_i)$ operation on channel \textit{i} is thus defined as,
$Y_i$(v)=Max({$X_i$($\bar{v}$): $\bar{v}$ in neighborhood of v}). In 3D, for example, the neighborhood can be a
3 x 3 x 3 cube around each voxel. The convolutional and max-pooling layers form the backbone of a CNN. A CNN architecture is constructed by combining multiple layers that successively learn more complex features from the input images. For example, with L layers the output can be mathematically expressed as ($O_{w(L)}$,...P $\circ$ $O_{w(1)}$)(X). Since we are considering an image classification problem, we add fully connected layers to the flattened output at the end of a CNN. 

Research in visual recognition has shown that fully connected feedforward architectures don’t scale well to full images. Instead, neural network architectures with local connectivity, such as CNNs, are much more suitable when dealing with high-dimensional images. The shared weights of the CNN architecture facilitate learning with fewer parameters.  
3D Convolutional layers thus transform an input 4D (3D multi-channel) volume to an output 4D volume. 
Each layer learns a set of spatial filters that activate in response to distinct visual patterns. Replicating or convolving each filter across the volume allows the corresponding pattern to be detected irrespective of its spatial location. Finally, the outputs from all filters are stacked along the 4th dimension to create a 4D feature map. Multiple convolutional layers coupled with pooling operations create global representations from local patterns. Stacking fully connected layers at the end after convolutional and down-sampling operations dramatically reduces the model parameter count for classification.

In our proposed approach, the input to the CNN is formed by concatenating voxel-level maps of ``connectivity fingerprints'', which are represented as a multi-channel 3D volume. 
Each channel is a connectivity feature, such as the Pearson correlation between each voxel's time series and the average signal within a target ROI.
In our implementation, we use both atlas-based and stochastic brain parcellation schemes to define target ROIs. The total number of input channels thus represents the number of ROIs used for creating voxel-level fingerprints. 
%In our experiments, we employed a variety of atlases, which define a specific parcellation of the brain into ROIs (see below for details).
%Each atlas consisted of between 110 and 400 ROIs, where a larger number of regions corresponded to a finer scale parcellation.
For each parcellation scheme (atlas-based or stochastic), we trained a separate model.
%, which we report performance values for.
%At each nodal scale, we also implemented an ensemble learning strategy, where the prediction was computed by taking a majority vote of the models.

\begin{figure}[h]
\centering\includegraphics[width=1.2\linewidth]{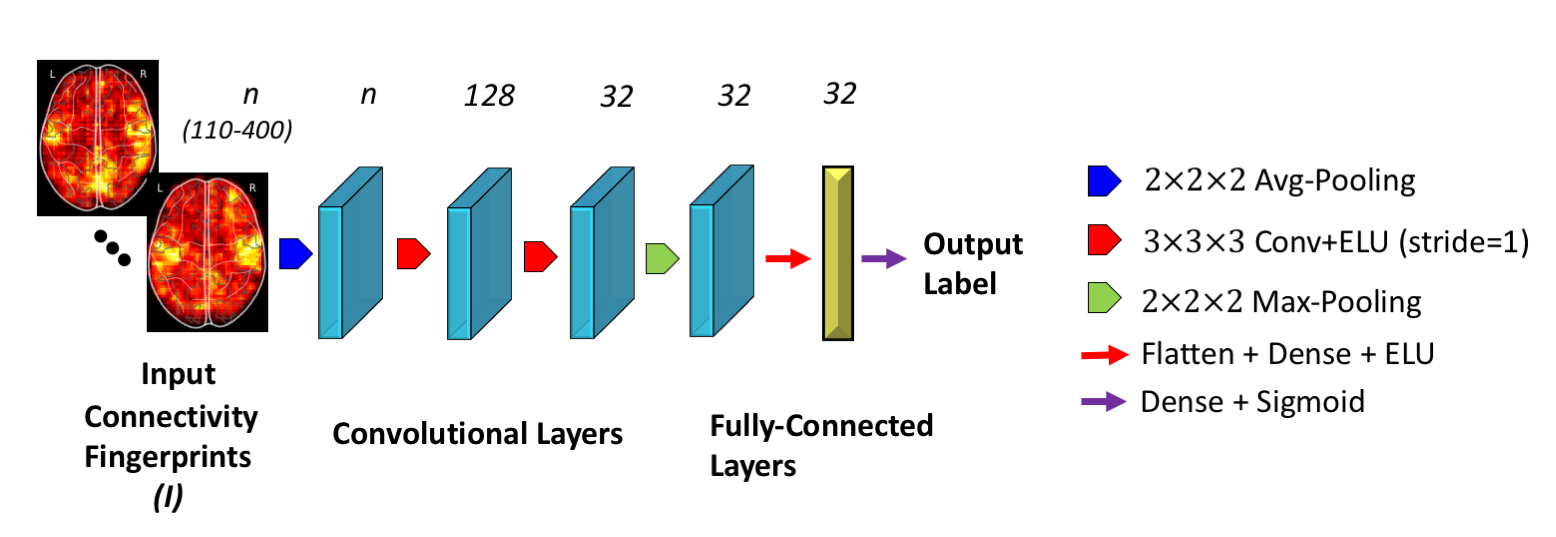}
\caption{Proposed CNN approach. All operations are in 3D volume. 2D correlation maps are shown for illustration only. For the age prediction task, an additional Max-Pooling and Batch-Normalization\cite{batchnorm} operation followed the first and second convolutional layer.} 
%\small{\footnote{Including an additional Max-Pooling operation following the first convolutional layer yielded better performance on the age prediction task.}}
\label{fig:conv_arch}
\end{figure}

In our experiments, we employed a simple CNN architecture, illustrated in Fig.~\ref{fig:conv_arch}.
Our architecture has several convolutional layers, interspersed with max-pooling based down-sampling layers, followed by a couple of densely connected layers.
The models were trained with a mini-batch size of 64, until convergence of validation loss. 
%For classification, the loss over the training set $D_{tr}\sim (x_i, y_i)$ was computed using binary cross entropy,
%\begin{equation}
%    \mathcal{L}(D_{tr})=\Sigma_{(x_i,y_i) \in D_{tr}}[y_i %\textrm{log}f(x_i,w)+(1-y_i)\textrm{log}(1-f(x_i,w))].
%\end{equation}
%For regression, we adopted standad squared loss
For classification, we used binary cross-entropy, whereas for regression we adopted mean squared difference as the loss function.
The neural network weights were optimized via stochastic gradient descent (SGD) for classification and Adam for regression. 
The learning rate and momentum for SGD were set to 0.001 and 0.9 respectively. Learning rate of Adam was set to 0.0005. For age regression, we employ a stochastic weight averaging strategy where we average the neural network weights over last 20 epochs.   
The same architecture and settings were used for all atlases and stochastic parcellations. 
We note that each atlas is defined on a unique gray matter mask. 
To ensure that all prediction models (benchmark and proposed) relied on information from the same voxels, the atlas-specific gray matter mask was applied to the voxel-level connectivity fingerprint data before feeding into the proposed convolutional architecture. 
For stochastic parcellations, the custom gray matter mask as described above was used for masking the fingerprints. 
The code and stochastic parcellations have been made available at: \url{https://github.com/mk2299/Ensemble3DCNN\_connectomes}.

\subsection{Benchmark Methods} \label{benchmark}
In our experiments, we implemented following benchmark methods.
\subsubsection{Ridge Regression} 
A linear regression model was trained with squared loss and $\alpha$ times the squared norm of the weight vector (See Appendix).
For classification, the ground truth labels were encoded as $\pm$ 1 for the two output categories.
We tested 10 linearly spaced values for the hyper-parameter $\alpha$ in the range [0.1,10] and report for the value with the highest cross-validation accuracy. 
%Thus this baseline result reflects an \textit{optimistic} estimate of performance.
%\vspace{-8pt}
\subsubsection{Support Vector Machine}
We implemented a standard SVM as a benchmark (See Appendix). 
We found that a radial basis function (RBF) kernel performed better than a linear model.
Thus we report results for the RBF-kernel SVM.
The two hyper-parameters (RBF kernel width $\gamma$ and  and misclassification cost weight $C$) were fine-tuned    by maximizing cross-validation accuracy via a grid search.
%Therefore, as with the ridge classifier, this should be considered as an upper bound on generalization performance.
For regression, we implemented the standard SVR scheme with an $\epsilon$- insensitive loss function, optimizing for the $\epsilon$-tube and penalty parameter of the error term via grid search. 
%CAN YOU INSERT A SENTENCE TO DESCRIBE THE SVR?

\subsubsection{Fully Connected Architecture}
%\newline 
The fully-connected neural network (FCN) architecture takes as input functional connectivity estimates between pairs of ROIs, which is vectorized and processed by a feed-forward network.
We implemented following architecture, which performed best on ABIDE-I cross-validation: 4 fully connected hidden layers, with 800, 500, 100 and 20 numbers of features and each linear layer followed by an elementwise Exponential Linear Unit (ELU) activation. Dropout regularization parameter was set to 0.2 and applied to each layer during training.
%The activation of node \textit{m} in layer \textit{l} (\textbf{$z_m^l$}) is computed using node activations in the previous layer through a non-linear function \textit{f} as $z_m^l=f(w_{l,m}^T z_{l-1}+b_{l,m})$. We choose a non-linear ELU activation for the hidden layers, which is given as f(x)=$e^x$-1 for x$<$0,  f(x)=x for x$\geq$0. 
For classification, the output node was a sigmoid, and cross-entropy loss was used.
For age prediction, the sigmoidal output was replaced with a linear activation and mean squared difference was used as the loss function.
%Similar to the 3D CNN, model hyperparameters for the fully-connected neural network were optimized on the functional connectivity data estimated using CC200 atlas. 
%The final model comprises 4 hidden layers with the number of hidden units set as [800,500,100,20]. 
The models were trained with a mini-batch size of 64, until convergence of validation loss.  

SGD was used as the optimizer with learning rate and momentum set to 0.01 and 0.9 respectively for classification. 
For age prediction, a smaller learning rate of 0.001 was used.
%We monitored training curves to ensure that all trained models had converged before terminating the optimization.

\subsubsection{BrainNet Convolutional Neural Networks}
BrainNet CNN, originally proposed in \cite{Kawahara2017}, utilizes specialized kernels to handle connectomic data. Their work described novel edge-to-edge, edge-to-node and node-to-graph convolutional layers that can potentially capture topological relationships between network edges. 
For BrainNet CNN, we implemented the following architecture that worked best on ABIDE-I cross-validation: 1 edge-to-node layer with 256 filters, followed by a node-to-graph layer with 128 output nodes and finally a dense layer with single output. A leaky ReLU non-linearity with alpha equal to 0.33 was applied to the output of each layer except the last layer. The activation of the last layer was set to linear and sigmoid for the regression and classification tasks, respectively. 
Dropout regularization with rate 0.2 was used for the edge-to-node layer. Similar to \cite{Kawahara2017}, Euclidean loss was minimized for age regression, whereas cross-entropy loss was used to optimize the classification models. The models were trained for 1000 iterations using SGD with momentum equal to 0.9. The learning rate was set to 0.0005 for age prediction and 0.008 for ASD/Healthy classification. 
The training curves were monitored for atlases to ensure convergence.  

\subsection{Ensemble Learning}
In our experiments, we explored two ensemble learning strategies.
The first one is what we call multi-atlas ensemble (or MA-Ensemble). 
MA-Ensemble averages the predictions of the models of a specific method (e.g., BrainNet CNN) computed using each one of the seven atlases. 
For classification, the final prediction is computed as the majority vote of the individual binary class predictions.
For regression, the ensemble prediction is simply the mean.
The second ensemble strategy (SP-Ensemble) averages across the models of a specific method computed using stochastic parcellations.
In our experiments, unless stated otherwise, we used 30 stochastic parcellations at each of the following four spatial scales: 110, 160, 200 and 400 ROIs. 
%For each scale, we created N=100 unique stochastic parcellations. 
These scales were chosen in accordance with existing atlases.
Thus the SP-Ensemble's prediction was computed based on fusing 120 ($30 \times 4$ scales) models. 
We also implemented single-scale SP-Ensemble models, which averaged over the 30 parcellations at the same spatial scale.

\subsection{Visualizing the CNN model}
In order to understand the connectivity features captured by the CNN model, we employed the saliency map approach of~\cite{SimonyanVZ13}. 
This visualization technique computes the gradient of the output prediction with respect to the input image voxel values, i.e., the 3D volume, using a single backward pass through the trained neural network. 
We then computed voxel-level saliency as the maximum absolute gradient value across all input channels corresponding to different target ROIs. 
More formally, consider an input image $\textbf{I}$, representing the connectivity fingerprints of \textit{V} voxels with \textit{R} ROI signals. The saliency weights $\textit{w}$ $\epsilon  \mathbb{R}^{V\times R}$ are computed by taking the absolute value of the gradient of neural network output \textit{O} with respect to the input image, i.e., $w=|\frac{\partial O}{\partial I}|$. In order to obtain the saliency at the voxel level $S$ $\epsilon$ $\mathbb{R}^{V}$, we take the maximum across all the ROIs, i.e., $S_{i}=\max_{1 \leq j \leq R}w_{ij}$. 
Finally, to visualize an ensemble model, we averaged the individual saliency maps that made up the ensemble.

\begin{table}[t]
\centering\includegraphics[width=0.8\linewidth]{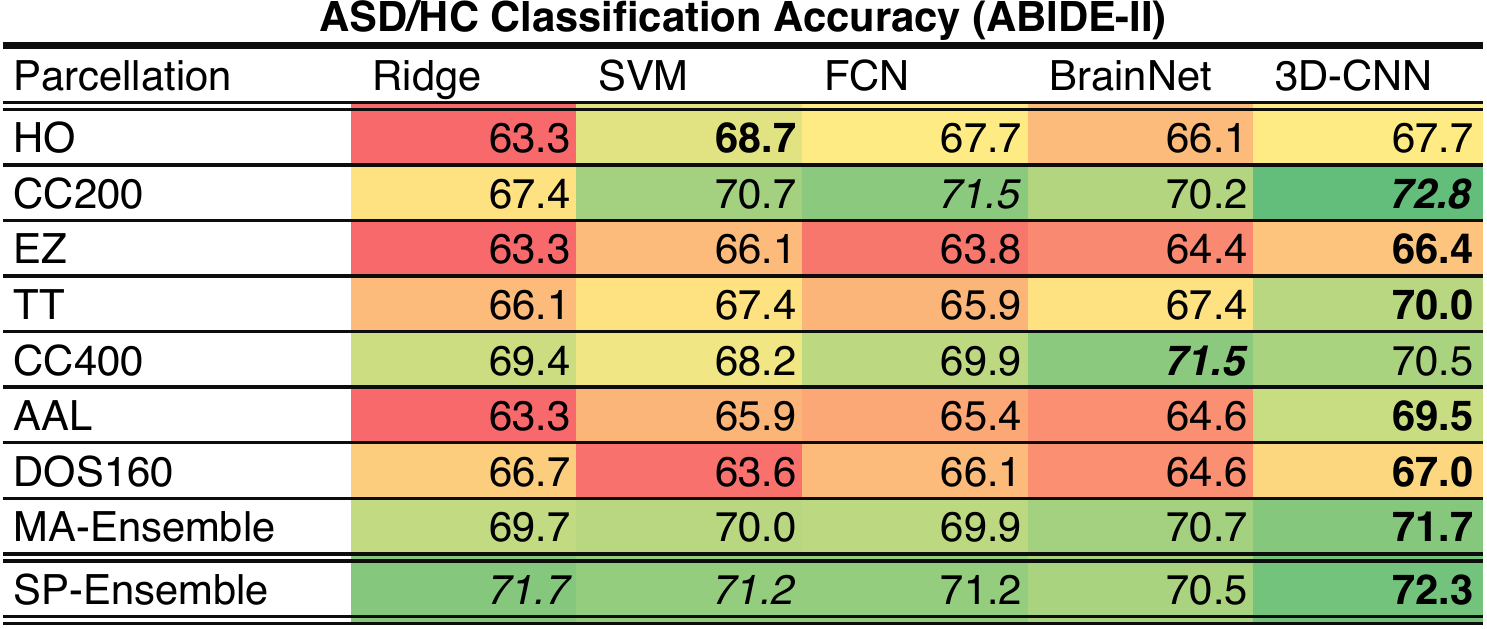}
\caption{Classification accuracy for ASD vs. Control: Independent test on ABIDE-II of baseline models and proposed CNN approach. For each row, best results are \textbf{bolded}. For each column, best results are \textit{italicized}. Green indicates better performance, whereas orange/red highlights worse performance.}
\label{tab:acc} 
%\label{fig:abide2dx}
\end{table}

\section{Results}
\subsection{Experiments}
In our experiments, we considered two tasks: i) binary classification of autism vs healthy, and ii) age prediction. 
For each task, we implemented two evaluation schemes. 
First, we conducted 10-fold cross-validation on the ABIDE-I dataset, so that we could present results that were comparable to previously reported classification results such as ~\cite{Plitt2015,ABRAHAM2017}.  
Second, we trained each model on the entire ABIDE-I dataset and computed test performance on the independent ABIDE-II set. 
%This is used for assessing the generalization behavior of different classifiers. 
We report classification accuracy and the receiver operating curves (ROC), along with corresponding area under the curves (AUC) for each of these scenarios under various combinations of parcellation schemes and prediction algorithms. 
For age prediction, we report the root mean squared error (RMSE). 

\begin{table}[t]
\centering\includegraphics[width=0.8\linewidth]{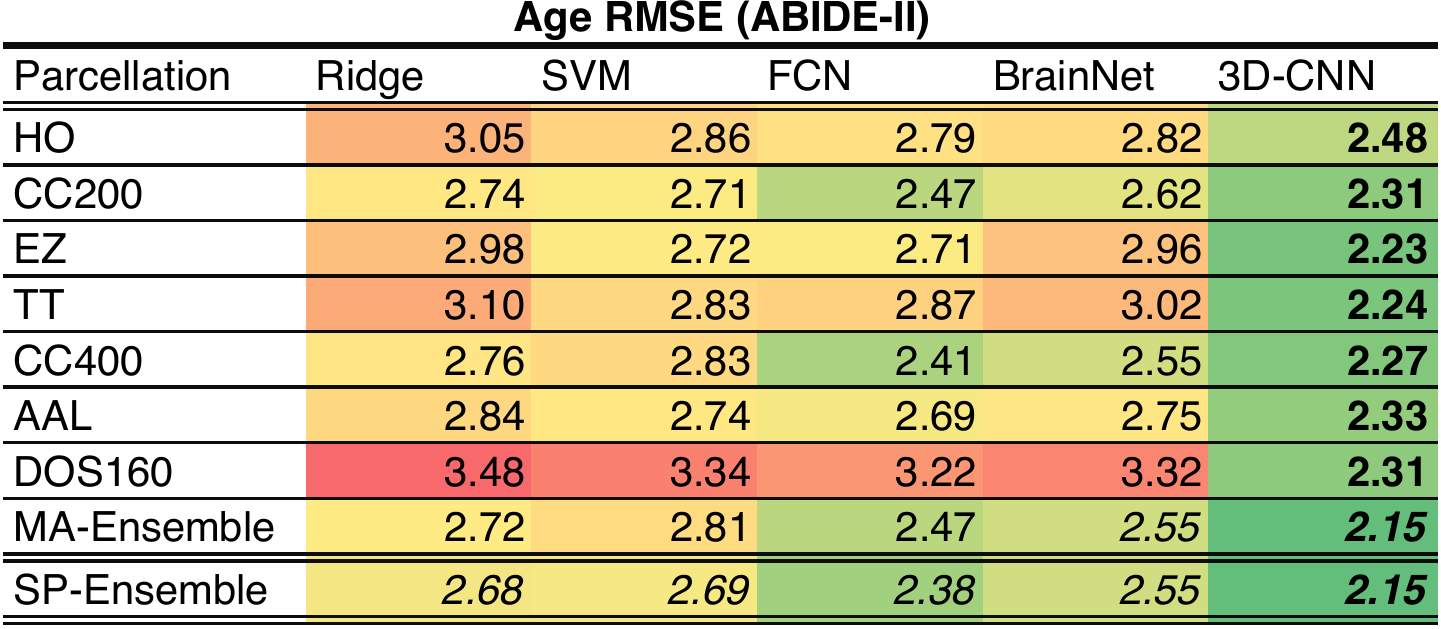}
\caption{Root mean squared error (RMSE in years) for age prediction: Independent test on ABIDE-II for benchmark models and proposed CNN approach. For each row, best results are \textbf{bolded}. For each column, best results are \textit{italicized}.}
\label{tab:mae}
\end{table}

\iffalse
\begin{table}[tbp]
  \centering 
  \hspace*{-0.5in}
  \begin{tabular}{l|l|l|l|l|l} %\hline
 % & \multicolumn{2}{c}{ABIDEI} 
  %& \multicolumn{2}{c}{ABIDEII-With Motion Scrubbing (QC)}  \\
 %  & ABIDE I & ABIDE II 
Parcellation &  Ridge & SVM & FCN& BrainNet & 3D-CNN \\ \hline
    HO & 2.60/2.39 & 2.90/2.40 & 2.62/2.28 & 2.64/2.31 & \textbf{2.54/2.25} \\ \hline
    CC200  & 2.54/2.22 & 2.90/2.26 & 2.64/2.22 & 2.59/\textbf{2.17} & \textbf{2.52}/2.20 \\ \hline
    EZ & 2.68/2.35 & \textit{2.87}/\textit{2.22} & 2.75/2.25 & 2.69/2.37 & \textbf{2.55/2.11} \\ \hline
    TT &  2.67/2.44 & 2.95/2.33 & 2.65/2.44  & 2.68/2.40 & \textbf{2.60/2.08} \\ \hline
    CC400  & 2.58/2.28  & 2.95/2.40 & \textbf{2.56}/2.19 & 2.59/\textbf{2.10} & 2.57/2.11 \\ \hline
    AAL &  2.72/2.26 & \textit{2.87}/2.25 & 2.83/2.39 & 2.76/2.19 & \textbf{2.48/2.07} \\ \hline
    DOS160 &  2.76/2.82 & 3.16/2.90 & 2.78/2.75 & 2.74/2.75 & \textbf{2.70/2.61}\\ \hline
    MA-Ensemble &  \textit{2.47}/2.22 & 2.90/2.34 & \textit{\textbf{2.42}}/2.22 & 2.45/2.14 & 2.43/\textbf{2.08} \\ \hline
    SP-Ensemble & 2.50/\textit{2.18} & 2.89/2.24 & \textit{2.42}/\textit{2.00} & \textit{2.43}/\textit{2.06} & \textbf{\textit{2.41}/\textit{1.94}} \\ \hline 
\end{tabular}
  \newline 
  \caption{Mean absolute error (MAE in years) for age prediction: 10-fold cross-validation on ABIDE-I healthy subjects/independent test on ABIDE-II for benchmark models and proposed CNN approach. For each row, best results are \textbf{bolded}. For each column, best results are \textit{italicized}.} 
  \label{tab:mae} 
\end{table}
\fi 

\begin{figure}[!ht]
%\begin{minipage}{0.5\textwidth}  
  %\centering 
  \hspace*{-1.5in}
 % \includegraphics[width=1.3\textwidth]{images/results_roc_abide1_cnn_edit.png} 
 % \end{minipage}
 \centering
\begin{minipage}{0.6\textwidth}  
  \centering 
  \includegraphics[width=1.5\textwidth]{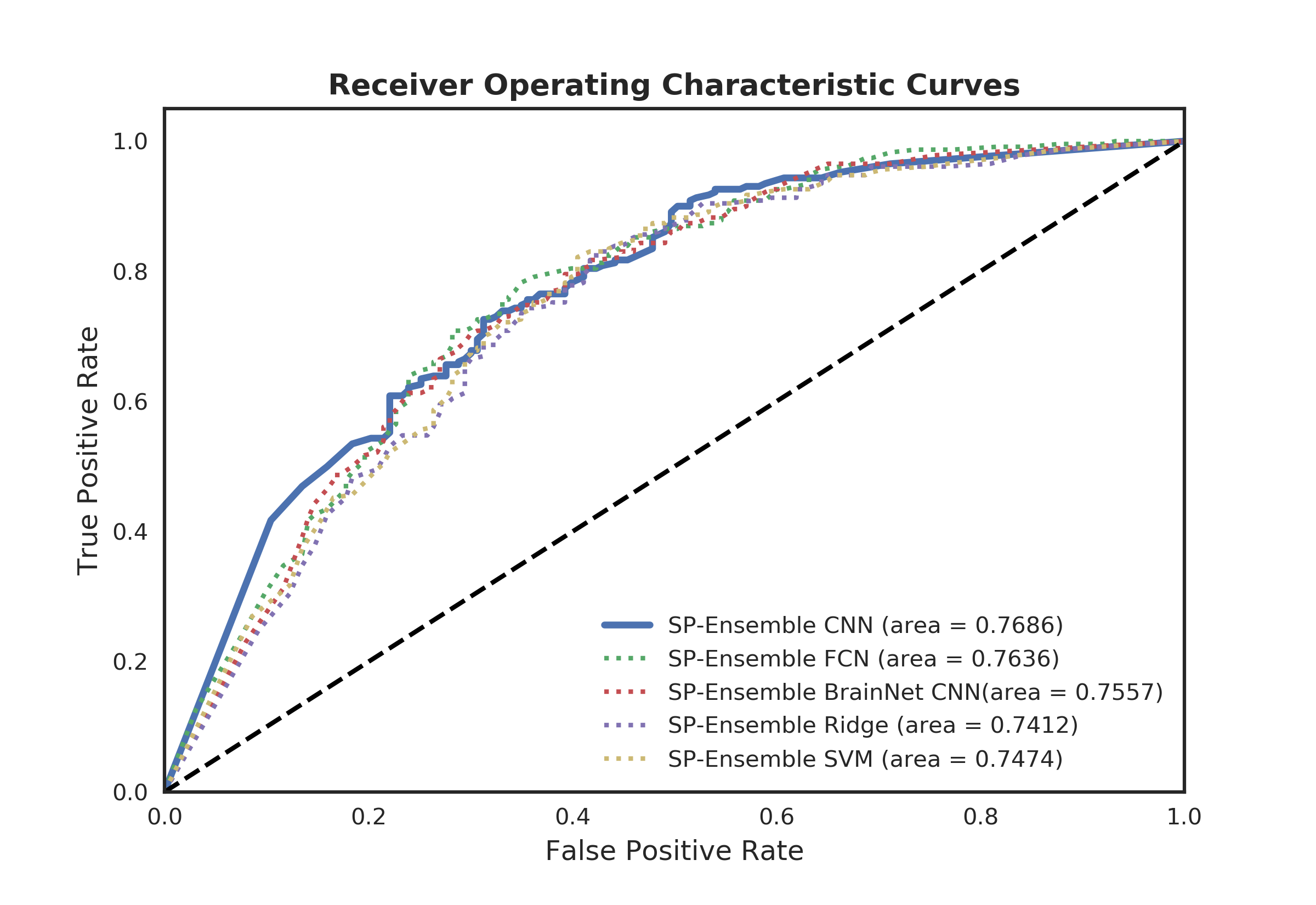} 
  
\end{minipage}
\caption{ASD-HC Classification: Receiver Operating Curves for independent validation on ABIDE-2}
\label{fig:roc}
\end{figure}

\subsection{Evaluation of Prediction Performance}
Table~\ref{tab:acc} shows the independent test performance for different models on the classification problem. 
The proposed 3D CNN approach performs at least as good as, and often better than, the benchmark methods, including the fully-connected deep neural network (FCN) and BrainNetCNN. 
In particular, the 3D CNN approach performs favorably against other algorithms for all but two parcellation schemes, including the ensembles.
Similarly, the SP-Ensemble achieves the best ABIDE-I cross-validation for most algorithms, including the 3D CNN.
The ABIDE-I cross-validation results, reported in Table~\ref{tab:abide1dx}, are in general compatible with the independent test results, where the 3D CNN and SP-Ensemble techniques mostly outperform the competition.
Figure \ref{fig:roc} shows the Receiver Operating Characteristic (ROC) curves for SP-Ensemble models for the different algorithms on the independent ABIDE-II test dataset. We observe that the 3D-CNN SP-Ensemble achieves an AUC of $\sim77\%$ and an accuracy of $\sim72\%$ on independent ABIDE-II data, slightly better than the state-of-the-art cross-validation on ABIDE-I for ASD/HC classification \cite{Heinsfeld18}, with FCN and Brain-Net CNN ensembles yielding a similar performance. ROC Curves for individual atlases are shown in Figure~\ref{fig:roc_atlases}. 

Table~\ref{tab:mae} lists independent test results for the age prediction task on ABIDE-II, and Table~\ref{tab:abide1age} reports the 10-fold cross-validation error on ABIDE-I.
The 3D CNN approach consistently shows superior performance, yielding the best results for all parcellation schemes. Similar to the classification scenario, SP-Ensemble or MA-Ensemble also yield the best cross-validation and independent test performance values for the majority of the algorithms, including 3D CNN. Overall, the best accuracy is achieved by SP-Ensemble 3D CNN, which yields a root mean squared error of \textbf{3.28} years on ABIDE-I cross-validation and \textbf{2.15} years on the independent ABIDE-II dataset. We also estimated mean absolute error (MAE) of all models on ABIDE-II and observed a similar trend, as reported in Table~\ref{tab:abide2maeage}. 
%It is important to note, however, that the input to the 3D-CNN model is different from the corresponding 

\begin{figure}
%\hspace*{-0.4in}
%\begin{minipage}{0.8\textwidth}  
\includegraphics[width=0.9\linewidth]{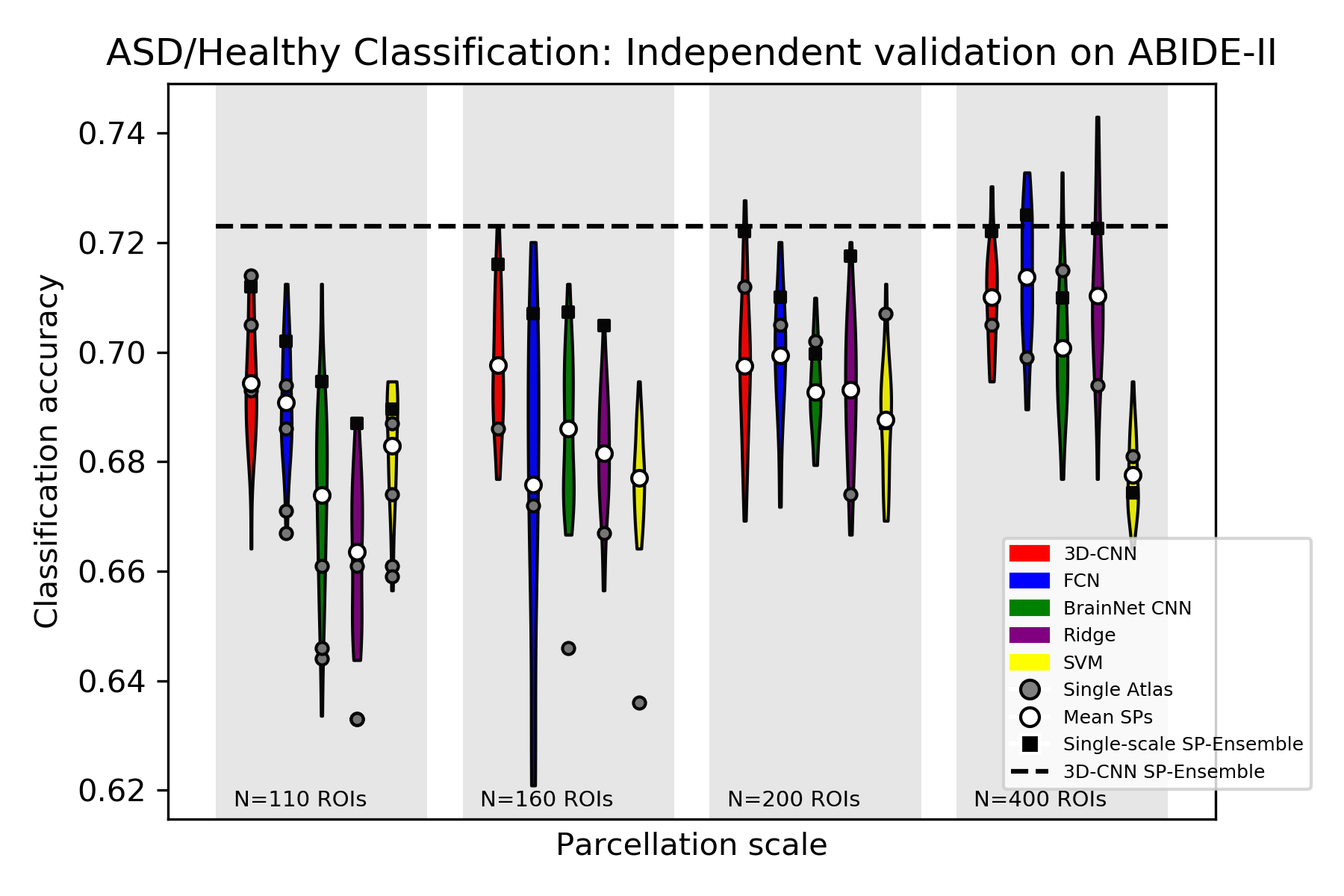}
%\caption{}
%\label{fig:cnn_violins_abide1_age}
%\end{minipage}
%\hspace*{-0.4in}
%\begin{minipage}{0.8\textwidth}  
\includegraphics[width=1.1\linewidth]{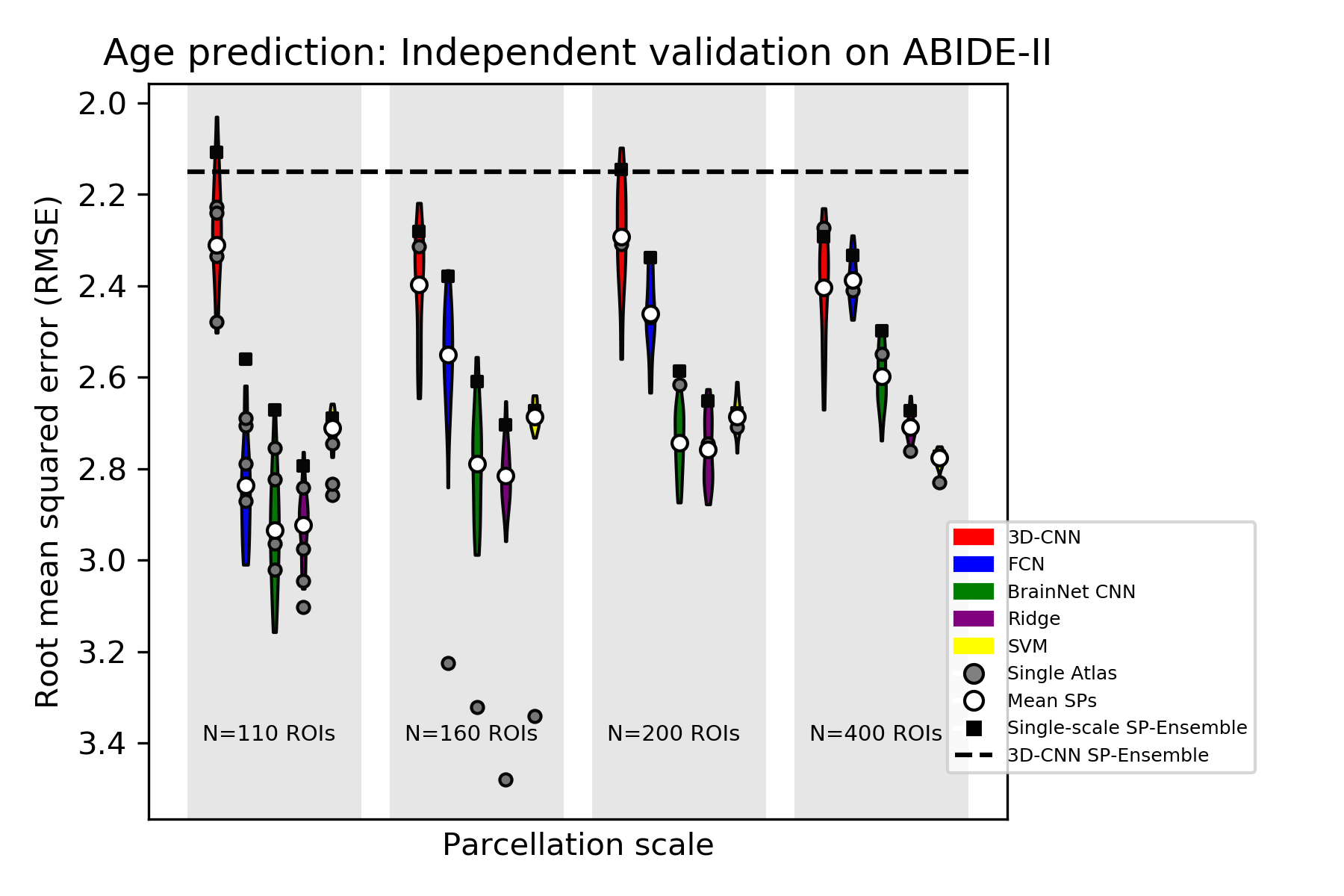}
%\caption{}
%\label{fig:cnn_violins_abide1_age}
%\end{minipage}
\caption{Violin plots showing the spread of prediction accuracies/errors for stochastic parcellations at multiple network scales for different classification models. Mean accuracy/error of individual violins is denoted by 'Mean SPs'. Performance of individual atlases is compared with SPs with the closest \# of ROIs and is denoted as 'Single Atlas'. Results are computed by training models on entire ABIDE-1 cohort and testing on the independent ABIDE-2 cohort.}
\label{fig:violins}
\end{figure}

\subsection{Comparison of stochastic parcellations and atlases}

Here, our objective is to conduct a detailed investigation of how the choice of ROIs affects prediction performance for different machine learning (ML) algorithms. 
%First, we generated 30 stochastic parcellations (See Appendix) at different nodal scales and trained the classifiers described in Sections \ref{3dcnn} and \ref{benchmark}, using the optimal hyperparameters for the best-performing atlas at that scale. 
For each ML algorithm and each parcellation we have a model trained on the ABIDE-I data, which we then used on the independent ABIDE-II data to quantify prediction accuracy. 
Figure~\ref{fig:violins} shows the distribution of accuracy values (estimated with a kernel density model) obtained using stochastic parcellations , while also illustrating the results for each of the atlases and the scale-specific SP-ensembles. 
The scale-specific SP-Ensemble strategy, as the name implies, averaged the models corresponding to the 30 stochastic parcellations in each scale. 
We observe that the atlas-based models performed no better than typical stochastic parcellation models, independent of scale and algorithm. 
This result offers an intriguing possibility:  perhaps we do not need anatomically or functionally derived brain parcellations to train machine learning models since stochastic parcellations perform equally well or no worse in practice. 
%This result presents a revelatory perspective: perhaps we do not need anatomically or functionally derived brain parcellations for machine learning.  

Our proposed SP-Ensemble CNN strategy yielded accuracy results that were about as good as the best scale-specific SP-Ensemble model.
Finally, the ensemble models were almost always better than the atlas-based models and they compared favorably against the individual stochastic parcellation models.
The same observations can be made for ABIDE-I cross-validation (see Supplementary Figure \ref{fig:violins2}). 

\begin{figure} %[h!]
\begin{minipage}{0.5\textwidth}  
  %\centering 
  \hspace*{-0.5in}
  \includegraphics[width=1.3\textwidth]{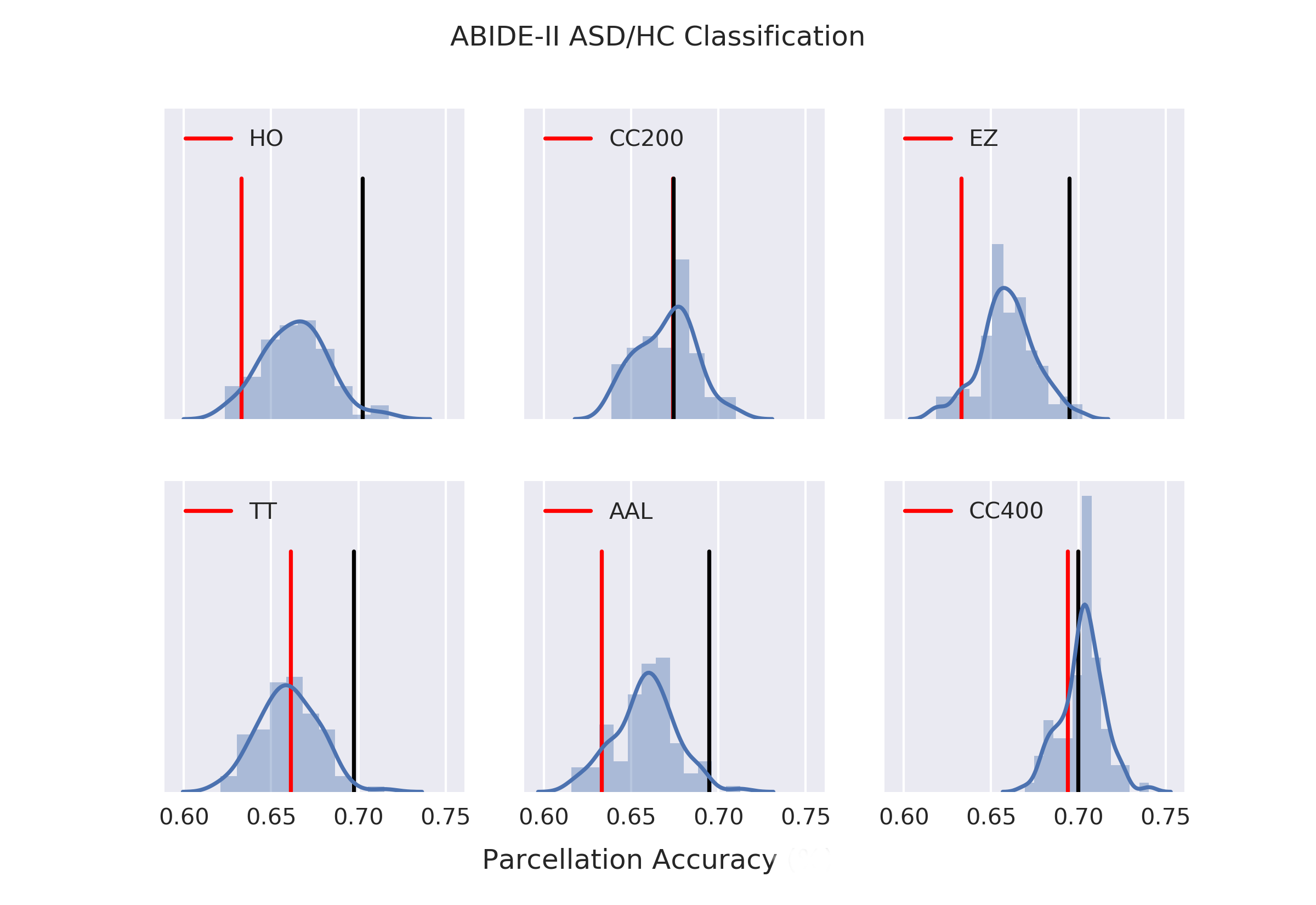} 
  \end{minipage}
\begin{minipage}{0.5\textwidth}  
  %\centering 
  \includegraphics[width=1.3\textwidth]{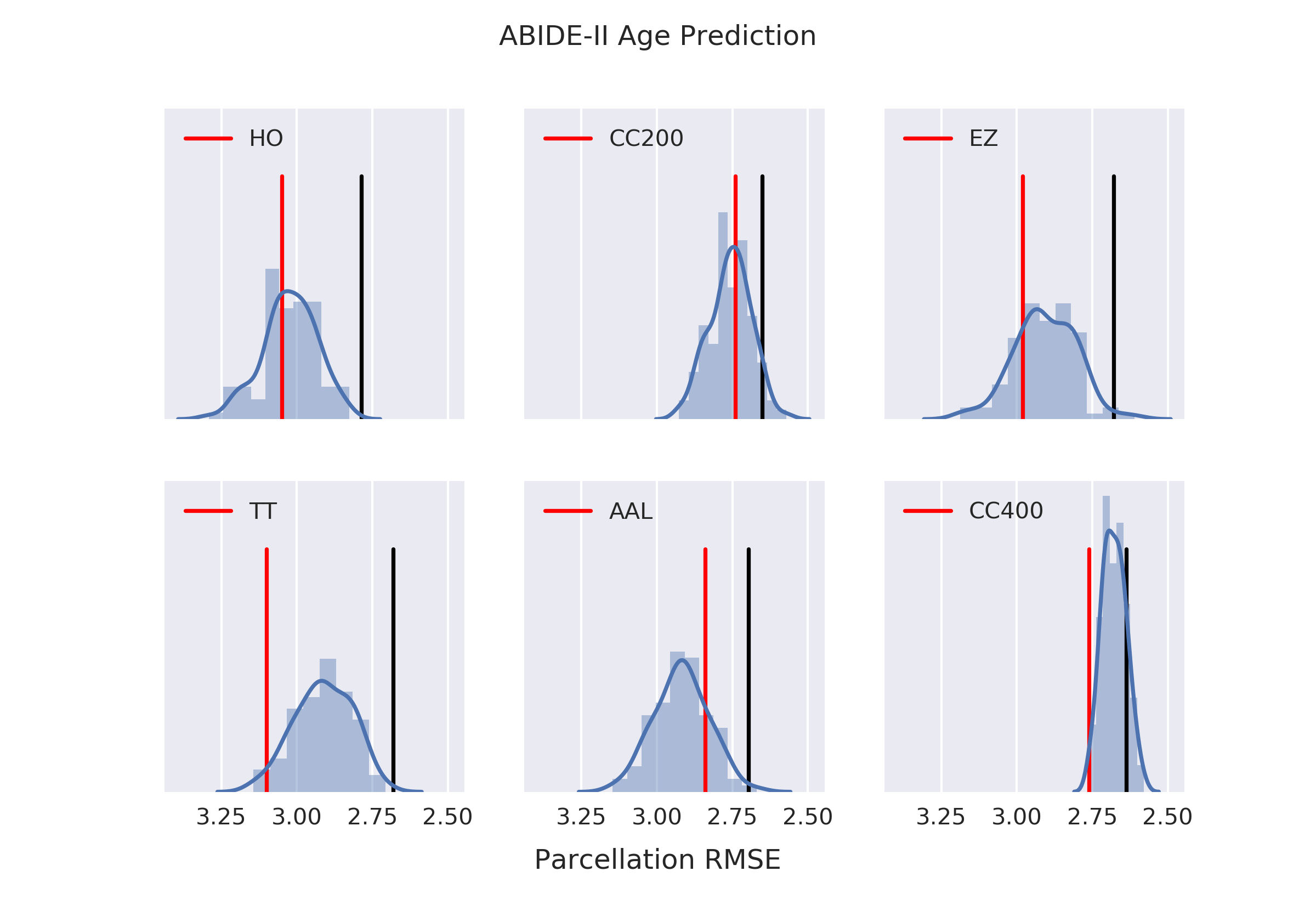} 
\end{minipage}
\caption{Distribution of Ridge models' performance for stochastic parcellations created using the same gray-matter mask as the corresponding atlas. Red denotes the atlas model's accuracy and black indicates the SP-Ensemble accuracy.}
\label{fig:sp100}
\end{figure}

In above analysis, one potential confound was the different gray matter masks of atlases and stochastic parcellations (SPs). 
In order to account for this confound, we conducted following analysis. 
For each of the atlases, we generated 100 SPs using the same gray matter mask as the atlas. 
We excluded DOS160 because it does not rely on a well-defined gray matter mask and places discontiguous 4.5 mm spherical regions over fixed coordinates in the brain (sampling only 5\% of brain voxels). 
We then trained on each of these SPs using the same hyper-parameters that were found to be optimal for the corresponding atlas. 
Here, we show the results for ridge regression (the model that was fastest to train), but we obtained similar results for all other algorithms as well.
%Because of the prohibitive computational costs associated with training the BrainNet and CNN architectures, this analysis was performed using the other benchmarks. 
As can be seen from Figure~\ref{fig:sp100}, for most atlases and corresponding gray matter masks, the model trained on the atlas ROIs performed no better than an average SP model.
Furthermore, and importantly, the SP-Ensemble (computed by averaging across SPs on the atlas-specific mask) yielded better performance than the atlas models for all atlases.

\begin{figure}
\begin{tabular}{@{}c@{}} %{0.8\textwidth}  
  \includegraphics[width=\textwidth]{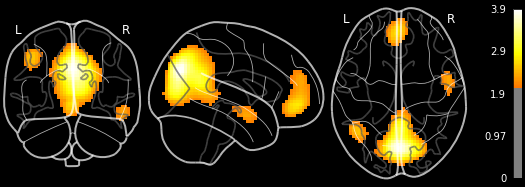} \\
   \small (a) ASD/Healthy Classification
  \end{tabular}
 
\begin{tabular}{@{}c@{}} %{0.8\textwidth}  
 % \centering 
  \includegraphics[width=\textwidth]{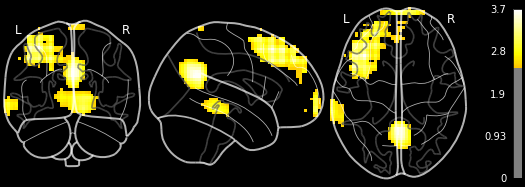}  \\
  \small (b) Age prediction
\end{tabular}

\caption{Mean saliency maps of trained 3D-CNN models for SP-Ensemble}
\label{fig:saliency_sp}
\end{figure} 

\subsection{Visualization}
An important goal of machine-learning tools in neuroimaging is to generate novel insights linking imaging biomarkers with disease or phenotypic traits. Visualization techniques for CNNs can help reveal important features used by the model for discriminating between output classes. Figure~\ref{fig:saliency_sp} shows the saliency maps computed for the SP-Ensemble CNN ASD classification and age prediction models. 
As can be seen from these maps, the precuneus, often considered a core node of the default mode network~\cite{utevsky2014precuneus}, seems to play a significant role for both prediction problems.
However, there are also salient regions that are unique to each problem.
For example, the anterior cingulate/ventromedial prefrontal cotex, a region that has been linked to autism~\cite{watanabe2012diminished}, was distinctly highlighted for the ASD classification problem. 
The left parietal cortex was also emphasized for ASD prediction, which is consistent with the laterilized activation observed in this region in Autism patients~\cite{koshino2005functional}.
On the other hand, for age prediction, the left dorsolateral prefrontal cortex (dlPFC) is a uniquely salient region.
The dlPFC is associated with executive functions, such as working memory and abstract reasoning.
For working memory, dlPFC's function seems to be age-associated and more lateralized in younger adults~\cite{reuter2000age}.

% \begin{figure} %[h!]
%   \hspace*{-0.5in}
%   \includegraphics[width=1.1\textwidth]{images/sp_atlas.png} 
%   \caption{Saliency maps of trained CNN models for different atlases}
%   \label{fig:saliency}
%   \end{figure}
  
%\begin{figure} %[h!]
%  \hspace*{-0.5in}
%  \includegraphics[width=1.1\textwidth]{images/sp_visualization.png} 
%   \caption{Saliency maps of trained CNN models for stochastic parcellations}
%   \label{fig:saliency_sp}
%  \end{figure}

\section{Discussion}
In this study, we presented a detailed empirical analysis of how the choice of ROIs can impact the performance of machine learning models trained on functional connectomes.
We considered several machine learning algorithms, together with a range of spatial scales and parcellation schemes, including the popular atlas-based techniques and a stochastic approach. %to report optimal learning strategies for connectomes that would likely generalize to other classification tasks as well. 
Our analysis suggests that using a single atlas for summarizing the connectome data is often sub-optimal for training machine learning models, and significantly more accurate predictions can be achieved with an ensemble approach that averages across models trained with different parcellation schemes. 
Furthermore, we demonstrated that averaging across stochastic parcellations can achieve very high accuracy values, often surpassing atlas-based models. Our findings resonate with several other studies that compare stochastic parcellations and atlases, although in different contexts. Craddock et al. \cite{Craddock2012} compared spatially constrained functional parcellations obtained from spectral clustering with anatomically constrained parcellations produced from random clustering. Random parcellations performed as well as functional parcellations and better than anatomical atlases on metrics of cluster homogeneity and representation accuracy. Based on this, the study reflected that sufficiently small ROIs perform well for functional network analysis regardless of their spatial position. Fornito et al. \cite{pmid20592949} generated stochastic parcellations by randomly sub-dividing the AAL atlas and showed that functional organizational properties are independent of the parcellation template at the same network resolution, although significant variability is observed across scales. Studies on diffusion-MRI based anatomical networks have similarly shown that topological attributes and network organizational parameters are consistent across different parcellation schemes, including random parcellations \cite{pmid20035887, Schirmer}.

% To the best of our knowledge, no previous study has investigated an ensemble learning approach across parcellations for building connectome-based classifiers, and we believe this is an important direction. 
% The Automated Anatomical Labelling and Harvard-Oxford atlases are thus far the most popular choices for studying functional connectivity \cite{Stanley2013}. However, as demonstrated here, these atlases show sub-optimal prediction performance by a significant margin. 

Another main contribution of this study is a novel approach to employ a 3D CNN architecture on functional connectivity data. 
Convolutional neural networks achieve state-of-the-art performance on many image-based prediction tasks, as they take advantage of the full spatial resolution of the data and the translation invariance property of the problem. 
Our proposed approach treats voxel-level connectivity fingerprints as input channels to a conventional 3D CNN framework. 
Spatial convolutions can capture local structural or topographic patterns in the data, such as connectivity gradients. Successively stacking convolutional layers in our architecture would hierarchically yield higher-order features that can capture information relevant for classification. Studies have shown that individual-level network topography serves as a fingerprint of human behavior \cite{pmid29878084}. Our multi-channel input image comprising connectivity fingerprints, coupled with CNNs, provides a natural framework to capture individual-level differences in topography as they relate to behavior or disease. This strategy contrasts with current practice where the input to machine learning models are pairwise ROI functional correlations. This makes the model more susceptible to uncertainty caused by parcellation choice. This can be seen in our experiments where there is relatively larger variance in prediction performance across atlases for the fully-connected neural network. Thus, CNNs with connectivity map inputs can offer a more robust alternative to classification approaches that only rely on ROI-level connectivity information, such as the BrainNet-CNN. Our results demonstrate that when tailored for connectomes, CNNs offer a promising opportunity to probe brain networks in disease. 

%This has important implications for connectome-based ML models, which can potentially benefit by employing ensembles of random parcellations. 

% From the perspective of interpretability, it would be interesting to compare the parcellations and understand why some of them perform better than others. Parcellating the brain randomly while establishing node correspondences can reveal how each data representation complements the other. It could further allow the design of efficient unified deep learning architectures with parameter sharing across distinct parcellations.  

Machine learning practitioners have to make a number of preprocessing choices in extracting connectomic features to analyze. While there is no one-size-fits-all solution across different tasks, in the context of machine learning models of functional connectivity, we present some interesting empirical observations below. 

\subsection{Ensemble learning}
The motivation behind using multiple stochastic parcellations for prediction is grounded in the concept of ensemble learning. 
The core idea is to integrate out a latent variable (i.e., parcels or ROI definitions) from the learning problem \cite{Mota2013a}. 
This approach also makes the predictions more robust to the precise parcellation scheme. 
As shown above, the performance of atlas-based models can vary significantly ($\sim$5-10\% for parcellations at the same scale). 
In such a scenario, ensemble learning over multiple stochastic parcellations can be a robust strategy that yields reliable predictions.  

\subsection{Network granularity}
We explored the impact of network granularity on prediction performance of machine learning algorithms for connectomes. Our analysis suggests that better prediction performance can be expected with parcellations at higher granularity upto $\sim $ 400 ROIs. To further investigate this trend on ROI-level models, we trained the fully-connected network (FCN), that is generally the best performing baseline algorithm, on both the prediction tasks for the 1024 node parcellation proposed in \cite{pmid20035887}. 
 As can be seen from Table \ref{tab:zalesky}, an atlas with 1024 regions is comparable to the CC200 atlas for ASD/HC classification in ABIDE-II. However, the performance actually degrades significantly (in comparison to CC200 or CC400) for the age prediction task. 

%However, no significant differences were observed beyond 400 ROIs. 

\begin{table} %[]
    \centering
   \includegraphics[width=0.6\textwidth]{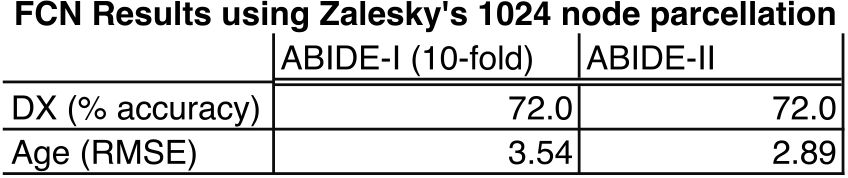}
    \caption{Classification/regression performance of FCN with a high-resolution parcellation ( $\sim$ 1024 ROIs) \cite{pmid20035887}}
    \label{tab:zalesky}
\end{table}
Our evaluations contradict with a previously reported result that a coarser network scale ($\sim$ 100-150 ROIs) is more suitable for autism classification~\cite{ABRAHAM2017}. 
In their paper, these conclusions were drawn by comparing the performances achieved with a few atlases. However, inferring trends from a small number of atlases can be misleading, since factors like the boundary definitions of structures (cortical/subcortical) or the particular gray matter mask used, will effect results. Stochastic parcellations can control for these confounds and depict unbiased trends across network scales.

% \subsection{More homogenous parcellations at higher resolutions}
% The narrower range in the prediction accuracies of stochastic parcellations at higher resolutions could be a result of increased functional homogeneity and less variability across parcellations. Consequently, the model becomes increasingly less sensitive to the parcellation scheme at higher resolutions. Further, this also explains why the CNN architecture demonstrates much less variability in prediction performance compared to other classifiers, since it is based on connectivity profiles at the voxel-level. 
% Smaller parcels are more likely to contain a single connectivity pattern, and this may in part also explain the higher accuracies obtained using finer parcellations.

\subsection{Number of gray matter voxels}
Our empirical study suggests that there is no direct correlation between the number of voxels in the gray matter mask and a model's prediction performance. 
However, we do observe that the choice of gray matter mask can impact results. 
%Mean accuracy of stochastic parcellations computed using our custom mask is consistently higher than the atlases at low resolution. This effect can also be observed across atlases. 
%KJ: I don't understand these numbers. The MNI 3mm atlas for DOS160 includes 3039 voxels (82cm^3), whereas the other atlases include ~20x more voxels
For example, the DOS160 atlas with as few as $\sim$ 3,039 voxels shows performance no worse than other atlases at the same resolution (HO, EZ, TT and AAL) with $\sim$ 20x more voxels.

\subsection{Visualization}
Saliency maps provide a valuable visualization strategy to probe deep neural network models. We visualized the saliency maps from 3D CNN models trained on ROIs extracted using both atlases and stochastic parcellations. As shown in Figure \ref{fig:saliency_sp} and Supplementary Figures \ref{fig:saliency_sp_individual} and \ref{fig:saliency_atlas}, these maps are remarkably consistent. 

These maps reveal that the precuneus, which is a hub of the default mode network and associated with ASD and age, plays an important role for both prediction problems.
There were also uniquely highlighted regions, such as the anterior cingulate/ventromedial prefrontal cortex for ASD classification and the left dorsolateral prefrontal cortex (dlPFC) for age prediction. 
Several studies have suggested the potential of DMN connectivity as a neurophenotype of autism. Chen at el. \cite{pmid26106547} trained a random forest classifier that distinguished ASD subjects from healthy controls with high accuracy, and showed that default mode and somatosensory regions contribute significantly to diagnostic accuracy. Similarly, Abraham et al. \cite{ABRAHAM2017} revealed discriminative connections in the DMN  for ASD/HC classification within a larger heterogeneous cohort of the ABIDE dataset. Furthermore, it has been shown that the connectivity of posterior cingulate cortex (PCC) and aberrations in the medial prefrontal cortex node of the DMN can predict social deficits in children with ASD \cite{pmid24183779}. Our results corroborate the findings of these studies, and suggest a crucial involvement of DMN in autism.

\subsection{Influence of motion} 
\begin{figure} %[h!]
%\begin{minipage}{0.5\textwidth}  
  \centering 
  \hspace*{-0.5in}
  \includegraphics[width=0.8\textwidth]{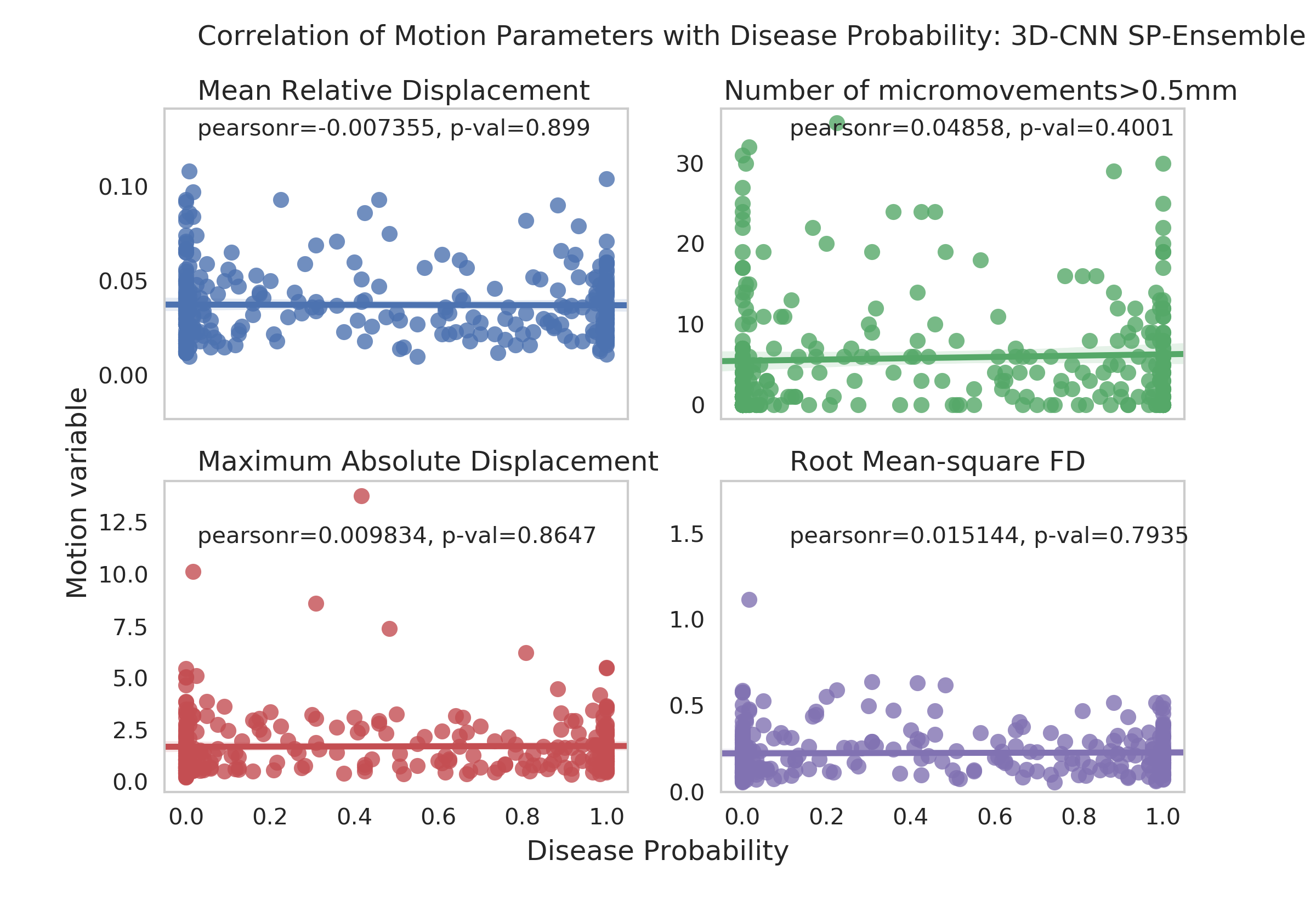} 
  %\end{minipage}
%\begin{minipage}{0.5\textwidth}  
  %\centering 
  %\includegraphics[width=1.3\textwidth]{images/fcn_motion_151sub.png} 
  
%\end{minipage}
\caption{Motion correlations}
\label{fig:motion}
\end{figure} 
Several studies have shown differences in head motion parameters during fMRI between healthy controls and diseased populations, or between subjects from different age groups~\cite{SATTERTHWAITE2012,Fair2012}. This, in turn, can manifest as artifacts in the derived resting-state connectivity~\cite{van2012influence}. Although our independent test data was motion scrubbed, we performed additional analyses to rule out the confounding effect of motion in classifier decisions. 
We selected a cohort of 151 ASD subjects with motion-matched healthy controls from our independent dataset and analyzed the correlation of 4 motion parameters with classifier predictions. 
These include the root-mean-square framewise displacement, mean relative displacement, maximum absolute displacement and the number of micro-movements greater than 0.5mm. These summary statistics were chosen in accordance with previous reports of motion artifacts in rs-fMRI\cite{Power2014}. As shown in Figure \ref{fig:motion}, no significant correlations were observed between motion variables and the predictions of SP-Ensemble (model average over all atlases). In this motion-matched cohort, classification accuracy of 71.8\% was obtained using 3D-CNN.

For our regression task, there was no significant correlation between a subject's age and any of these motion parameters in our cohorts. 
 
 \subsection{Recommendations}
\begin{figure} %[h!]
\begin{minipage}{0.5\textwidth}  
  %\centering 
  \hspace*{-0.5in}
  \includegraphics[width=1.3\textwidth]{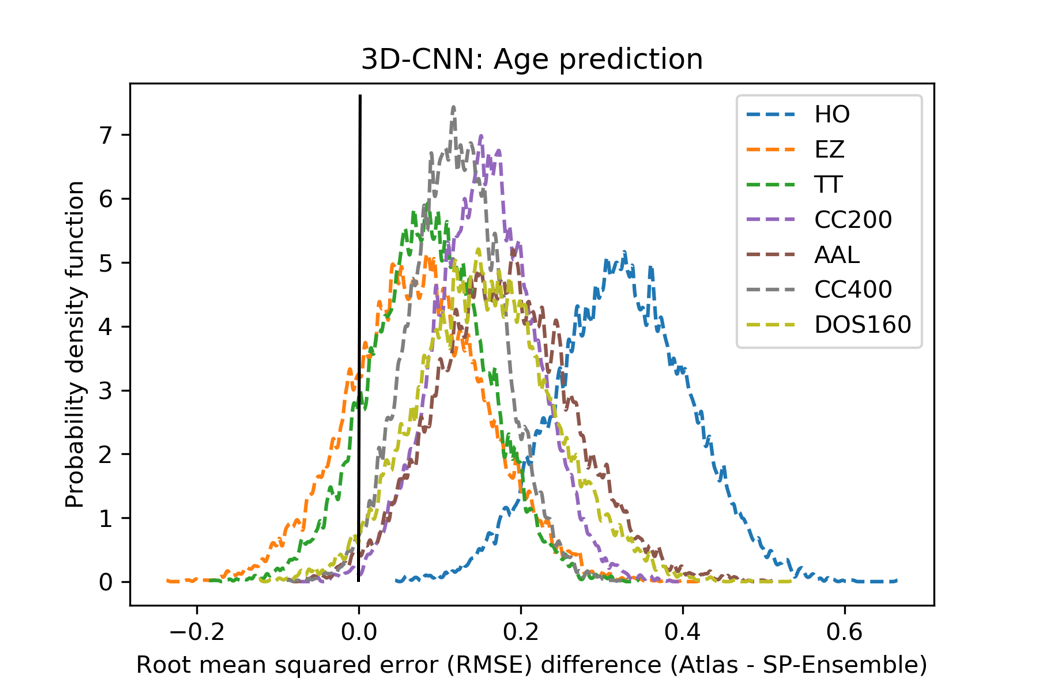} 
  \end{minipage}
\begin{minipage}{0.5\textwidth}  
  %\centering 
  \includegraphics[width=1.3\textwidth]{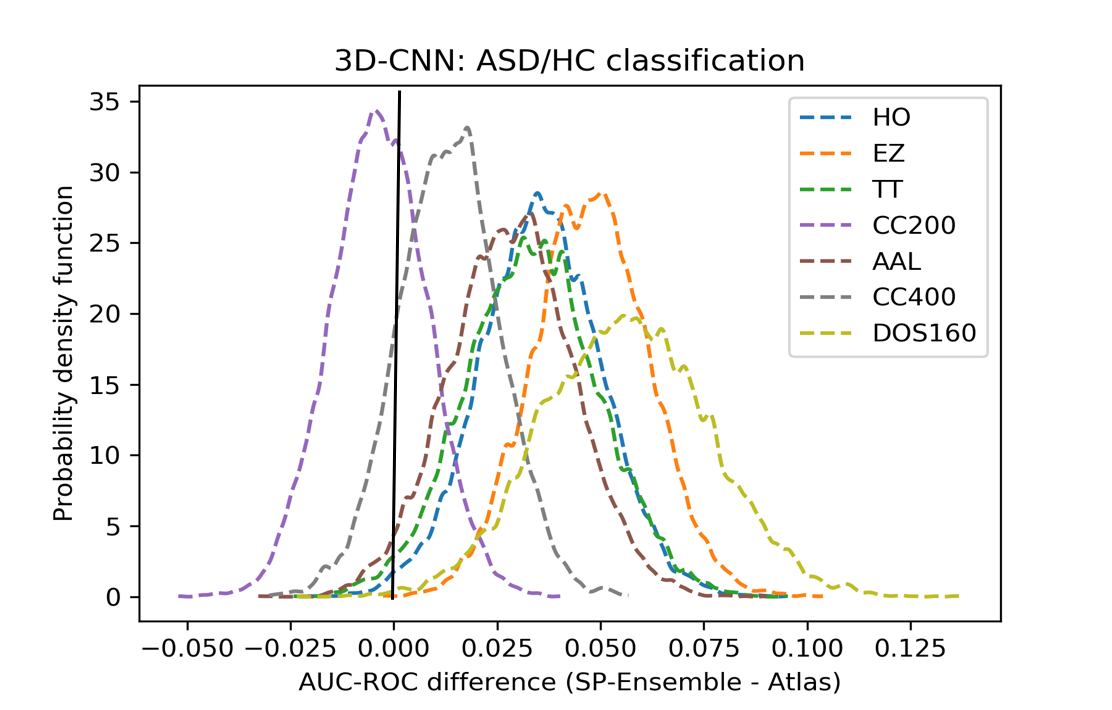} 
\end{minipage}
\begin{minipage}{0.5\textwidth}  
  %\centering 
  \hspace*{-0.5in}
  \includegraphics[width=1.3\textwidth]{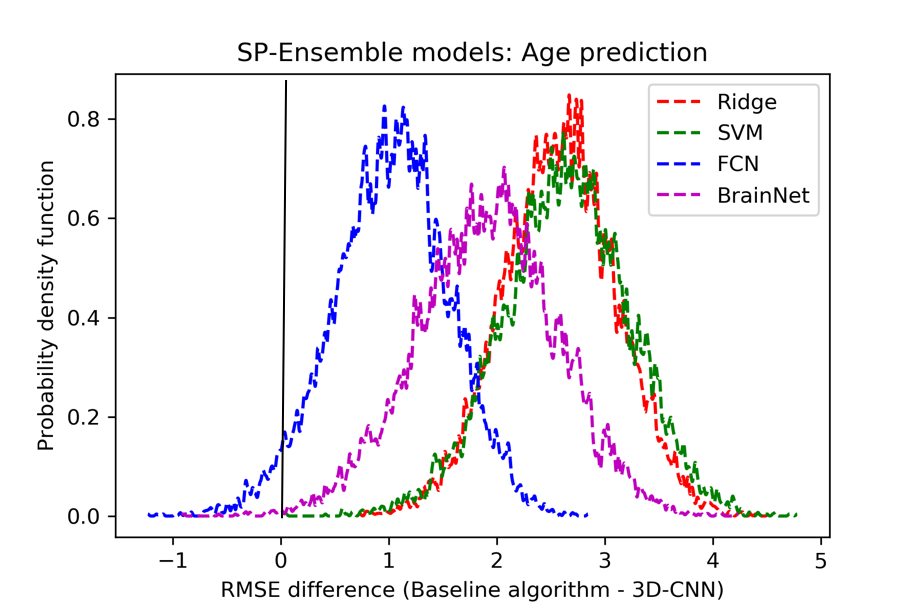} 
  \end{minipage}
\begin{minipage}{0.5\textwidth}  
  %\centering 
  \includegraphics[width=1.3\textwidth]{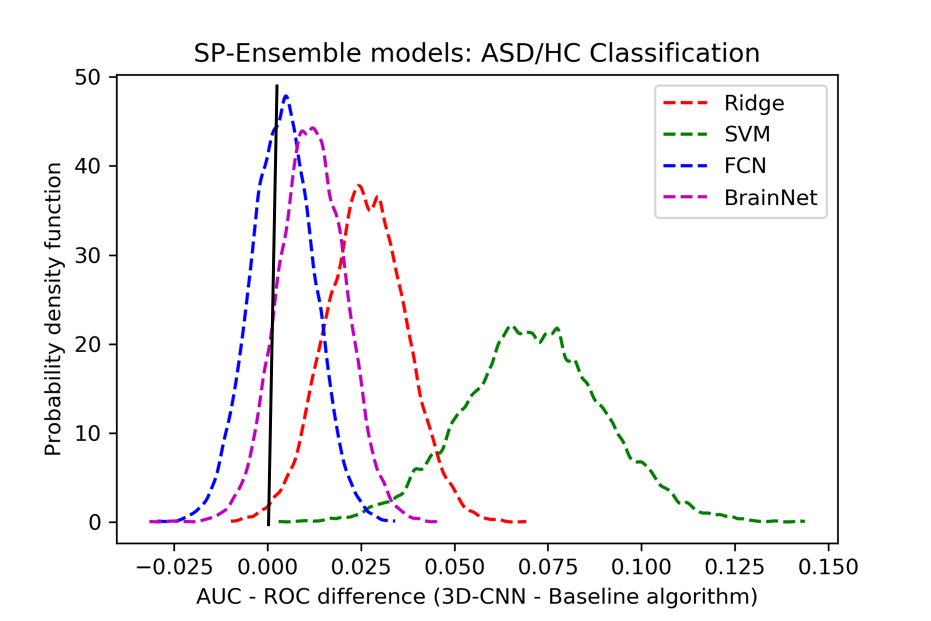} 
\end{minipage}
\caption{Kernel density estimates of the probability distributions for the performance difference between models, computed based on 10000 bootstrap samples from ABIDE-II. Values to the left of the black vertical line indicate bootstrap samples where the proposed approach (3D CNN or SP-Ensemble) under-performed compared to the competing method. }
\label{fig:kde}
\end{figure}

 Based on our experiments, we make two claims in this study: (a) 3D-CNN performs favorably compared to alternative baseline algorithms, and (b) Ensemble models that average across parcellation schemes consistently perform better than individual atlas-based models and are thus a safer choice for supervised machine learning on connectomes. This is because individual atlases can show significant variability in classification/regression performance and finding the optimal atlas for a prediction task among the wide range of available atlases might not be feasible. 
Figure \ref{fig:kde} shows the probability density estimates for the difference in performance between (a) 3D-CNN versus baseline algorithms as evaluated with the SP-Ensemble strategy, and (b) SP-Ensemble versus single atlas implemented with the 3D-CNN model. These estimates are presented for both our prediction tasks. For this experiment, we estimate the evaluation metrics (AUC-ROC for ASD/HC classification and RMSE for age regression) on 10,000 bootstrapped samples from ABIDE-II. These results demonstrate that the SP-Ensemble approach consistently achieves an accuracy as good as the best performing single-atlas model. Further, the 3D-CNN model consistently outperforms the baseline algorithms for the age prediction task, with more prominent improvements for individual atlas models. This can be seen from Tables \ref{tab:acc} and \ref{tab:mae}. We note that when using the ensemble strategy, the differences between models are marginal and might be irrelevant in some practical applications. For instance, the SP-Ensemble performance on ASD/HC classification task is comparable among 3D-CNN, FCN or BrainNet-CNN, with slight improvements over linear models. Thus, if time and/or computational resources impose constraints, it might be more suitable to prefer simpler models like FCN or SVM over 3D-CNN for example, especially with the ensemble approach.    

\subsection{Limitations and future work}

Throughout our analysis, Pearson's correlation was chosen to measure functional connectivity strength between different brain regions. Several other correlation metrics, including tangent-based and partial correlation have been shown to yield superior classification performance in prior studies \cite{dadi2018,ABRAHAM2017}. While we do not expect this to affect the general conclusions and findings of our study, the choice of the correlation metric still remains an arbitrary decision in any machine learning pipeline for connectomes. 

Due to the heavy computational burden required for training multiple deep learning models, we only considered one particular scheme for creating stochastic parcellations, i.e., Poisson Disk Sampling. Alternative strategies for creating random parcellations have also been proposed, for instance, through stochastic sub-division of anatomically derived ROIs into smaller parcels \cite{Hagmann2008}. 
It is also possible to randomize several other more popular schemes for parcellating the brain, such as, using Ward's clustering on functional data from sub-samples of the population \cite{Mota2013a} or creating Geometric parcellations with different initializations \cite{ARSLAN20185}. 

While the proposed CNN approach achieves promising accuracy on autism detection and age prediction, there is room for further improvement. 
We have not yet conducted a comprehensive optimization of the convolutional architecture. Furthermore, there are likely more optimal choices than target ROI-based correlations that are used as input to the model. An interesting alternative would be select random gray matter vertices for connectivity profiling, as proposed in \cite{Yeo2011}. We envision an end-to-end learning strategy that can enable the optimization of these connectomic features.

Saliency maps provide an appealing visualization technique by mapping the neural network activations back to input voxel space. Several modifications to gradient-based back-propagation have been reported in literature that can potentially highlight more informative features learnt by the model \cite{zintgraf,Selvaraju16}. Further, the use of saliency maps need not be restricted to depicting group-averaged discriminative features. 
Unsupervised learning on saliency maps can provide novel insights into clinical subtypes of disease. 
It is also important to note that machine learning techniques do not unequivocally provide evidence for the salient features being directly associated with the disease or other target variables. 
However, when combined with detailed future investigations, they can spur clinical discoveries.

\subsection{Conclusion}
The results presented in our paper showcase the utility of ensemble learning for connectomes. 
Functional network based prediction models are impacted by several a priori choices, the most pivotal of which is the ROI definition. 
We demonstrate that ensembles of stochastic parcellations yield predictions that are significantly more robust and accurate compared to single atlas-based approaches. 
Further, our experiments highlight the potential of  convolutional neural network models for connectome-based classification. 
% Through modern visualization tools and saliency mapping, these prediction models hold significant promise for interpretable diagnostics or decision-support systems. With the public release of several large-scale datasets alongside open-source initiatives, the fMRI community is now accessing datasets with several hundreds or thousand subjects. With this continuous increase in the size of connectome datasets, neural-network based approaches can be made even more effective. 
%When tailored for connectomes, modern DNN architectures like Convolutional Neural Networks, offer an unparalleled opportunity to probe brain networks for individual-level predictions. 

\section{Acknowledgements}
This work was supported by NIH grants R01LM012719 (MS), R01AG053949 (MS), R21NS10463401 (AK), R01NS10264601A1 (AK), the NSF NeuroNex grant 1707312 (MS) and Anna-Maria and Stephen Kellen Foundation Junior Faculty Fellowship (AK).
\section{References}

%% References


\bibliographystyle{model1-num-names}
\bibliography{connectome.bib}






% \end{thebibliography}
\clearpage
\renewcommand\thefigure{S.\arabic{figure}}  
\renewcommand\thetable{S.\arabic{table}} 
\section{Supplementary Material}
\setcounter{figure}{0}
\setcounter{table}{0}
%\subsection{How many stochastic parcellations do we need?}
%Since there are innumerable ways of creating stochastic parcellations, the optimal number of parcellations to build the ensemble at each nodal scale was determined using a preliminary analysis. We used the fully-connected neural network architecture for this analysis since it offered a good compromise between classifier performance and computational burden during training. Figure \ref{fig:rois_acc} shows the accuracy (and the associated uncertainty) obtained using different number of parcellations in the ensemble (x-axis) at each network scale. Based upon this, we selected N=30 random parcellations for our ensemble framework. 
%Further, if we consider parcellations at the same scale, the effect of model averaging is more significant at lower resolutions. Beyond 200 network nodes, we observed that the accuracy plateaus very quickly, as shown in the figure. This is expected since there is low variability in parcellations at finer resolutions. 
 
%\begin{figure} %[h!]
%  \hspace*{-0.5in}
%  \includegraphics[width=1.1\textwidth]{images/nparcel.png} 
%  \label{fig:rois_acc}
%  \caption{10-fold cross-validated results on ABIDE-I}
%  \end{figure}

\subsection{Atlas Summary}
\begin{table}[!htbp]
 \begin{tabular}{||c c c c c c||} 
 \hline
Atlas & $\#$ of ROIs & Total Vol. & Median Vol.($\pm$ std) & Min Vol. & Max Vol.\\ [0.5ex] 
 \hline\hline
 TT & 97 & 1656.34 & 12.5 ($\pm 16.02$) & 0.03 &  69.71 \\ 
 \hline
 HO & 111 & 1611.39 & 10.04 ($\pm 15.26$) & 0.05 & 97.33 \\
 \hline
 EZ & 116 & 1941.65 & 14.11 ($\pm 11.96$) & 0.97 & 56.35 \\
 \hline
 AAL & 116 & 1843.10 & 13.78 ($\pm 11.05$) & 1.35 & 53.33  \\
 \hline
 DOS160 & 160 & 82.05 & 0.51 ($\pm 0.04$) & 0.03 & 0.51  \\
 \hline
 CC200 & 200 & 1172.15 & 5.83 ($\pm 1.26$) & 1.81 & 9.96  \\
 \hline
 CC400 & 400 & 1172.15 & 2.97 ($\pm 0.68$) & 0.76 & 5.35  \\
 \hline
\end{tabular}
\caption{Summary descriptors of ROIs in individual atlases. All volumes are in $cm^3$. } 
\label{table:atlas}
\end{table}

\subsection{Poisson Disk Sampling}\label{ssec:pds}
Poisson disk sampling is a stochastic sampling procedure where drawn samples are required to be at least a distance \textit{d} apart for some user-specific distance metric and density parameter. Since we use this sampling procedure to draw parcel centers, \textit{d} is estimated a priori based upon the desired number of parcels, and spatial proximity is used to compute the distance. We use the fast poisson disk sampling algorithm as proposed in \cite{pds}. This is an efficient sampling procedure that generalizes to arbitrary dimensions and allows volumetric sampling. 
The algorithm is outlined below:
\begin{itemize}
    \item \textbf{Step1:} A parcel center is arbitrarily chosen from all gray matter voxels and stored in an initially empty ‘active’ list. 
    \item \textbf{Step 2:} A sample \textit{c} is drawn from this active list of voxels. The next candidate parcel center is randomly selected from the list of all voxels within a spherical annulus between radius \textit{d} and \textit{2d} around c. A candidate is accepted and added to the active list if it is atleast a distance \textit{d} apart from all the existing parcel centers; otherwise another candidate is chosen. If no candidate is accepted from the annulus, c is removed from the active list and another sample \textit{c} is drawn. This procedure is repeated until the active list is empty. 
\item \textbf{Step 3:} Once the centers are sampled, every gray matter voxel is assigned to its closest parcel center. Sampling is performed for the left and right hemispheres separately to avoid parcels that cross hemispheric boundaries.
\end{itemize}

%\subsection{Parcellation masks}

\subsection{Linear Classifiers}
\subsubsection{Ridge Classifier}
Given feature vectors \textbf{$x_i$} for \textit{n} subjects and the corresponding prediction variables denoted by \textbf{$y_i$}, we approximate the fit using a linear regression model. An $L_2$ regularization for the weights (\textbf{w}) is added to the mean squared error to yield the following loss function of ridge regression:
\begin{equation}
 \mathcal{L_{R}}=\left\Vert Xw-y\right\Vert^2 +\alpha \left\Vert w \right\Vert^2 
\end{equation}
During classification, the output labels \textbf{y} are encoded as $\pm$ 1 for the two output categories to minimize the above loss.

\subsubsection{Support Vector Machines}
(a) Classification \\
Support Vector Machine Classifiers optimize for a hyperplane with maximum margin between the output classes. This results in a decision function of the form, \textit{f}(x)= \textbf{sign($w^Tx+b$)}. The weights \{\textbf{w}, b\} are obtained by minimizing the following convex loss function consisting of a data loss component ($\mathcal{L_{D}}$) and a regularization loss for the weights ($\mathcal{L_{W}}$),
\begin{equation}
 \mathcal{L_{SVC}}= C\mathcal{L_{D}}+\mathcal{L_{W}}%\left\Vert Xw-y\right\Vert^2 +\alpha \left\Vert w \right\Vert^2 
\end{equation}
$\mathcal{L_{D}}$ is modeled using a hinge loss function, $\sum_{i=1}^n$max$(0,1-y_i (w^Tx_i+b))$ over all \textit{n} training samples \{($x_1$,$y_1$),...,($x_n$,$y_n$)\}. $\mathcal{L_{W}}$ is modeled using a Euclidean norm, i.e., $\left\Vert w \right\Vert^2 $. 
Here, C is a tuning parameter that controls the trade-off between regularization and data loss. 
\newline
(b) Regression \\
The $\epsilon$-Support Vector Regression (SVR) scheme optimizes for a decision function of the form, \textit{f}(x)=$w^Tx+b$, that has at most $\epsilon$ deviation from the true prediction variables \textit{y} (allowing for errors when the problem is infeasible). The loss function ($\mathcal{L_{SVR}}$) can be formulated as,
\begin{equation}
 \mathcal{L_{SVR}}= C\mathcal{L_{\epsilon}}+\mathcal{L_{W}}
\end{equation}
$\mathcal{L_{\epsilon}}$ is traditionally referred to as the $\epsilon$-insensitive loss function, and is formulated as $\sum_{i=1}^n$max$(0,|w^Tx_i+b-y_i|-\epsilon)$ over all \textit{n} training samples \{($x_1$,$y_1$),...,($x_n$,$y_n$)\}. The regularization term ($\mathcal{L_{W}}$) is modeled using a Euclidean norm, i.e., $\left\Vert w \right\Vert^2 $. The tuning parameter C controls the trade-off between the regularization (i.e., the flatness of the decision function) and the amount up to which deviations beyond $\epsilon$ are tolerated. 
\newline
Both the classification and regression problems yield weights \textbf{w} that can be represented completely as a linear combination of the training inputs $x_i$. Thus, \textbf{w} is represented as $\sum_{i=1}^n\alpha_ix_i$, and the decision function becomes \textit{f}(x)=$\sum_{i=1}^n\alpha_ix_i^Tx+b$. This makes it easier to extend SVMs for non-linear decision functions using the kernel technique, i.e., by applying transformations $\phi$(x) that map x to a high-dimensional space and replacing the inner product $\langle x_i, x\rangle$ with the kernel $K(x_i,x)$=$\langle \phi(x_i), \phi(x)\rangle$. For our experiments, we observed that the radial basis function kernel, $K(x_i,x)=$exp(-$\frac{\left\Vert x_i -x\right\Vert^2}{2\sigma^2})$, yields the best results among linear, sigmoid and polynomial kernels up to degree 4.  

\subsection{Neural network hyperparameter settings}
We note that since it is more expensive to train a 3D-CNN, we could experiment with only a limited configuration of hyperparameters during cross-validation on ABIDE-I data, compared to FCN and Brain-Net CNN, which work with vectorized connectivity matrices and are thus faster to train. For all three neural network models, we relied on a random search over the learning rate, number of layers, number of units or feature maps in each layer and the choice of non-linearity. For this search we employed the HO atlas. The hyper-parameter configuration that yielded the best ABIDE-I cross-validation accuracy was subsequently used for all other parcellation schemes. Note that except for some minor changes, the models for age prediction and ASD/HC classification are almost identical. Furthermore, in our primary analyses we compare models based on ABIDE-II performance, which was not used for hyper-parameter tuning. 

For the FCN, we initially started with the architecture proposed by Heinsfeld et al.\cite{Heinsfeld18} for ASD/HC classification. We increased the number of layers before the softmax output until the ABIDE-I cross-validation accuracy stopped improving. Also, we noticed that adding batch-normalization after each layer had no noticeable impact on classification performance. Hence, we didn’t include this layer in our FCN architecture.

\subsection{ABIDE-I cross-validation results}

\begin{table}[t]
\centering\includegraphics[width=0.8\linewidth]{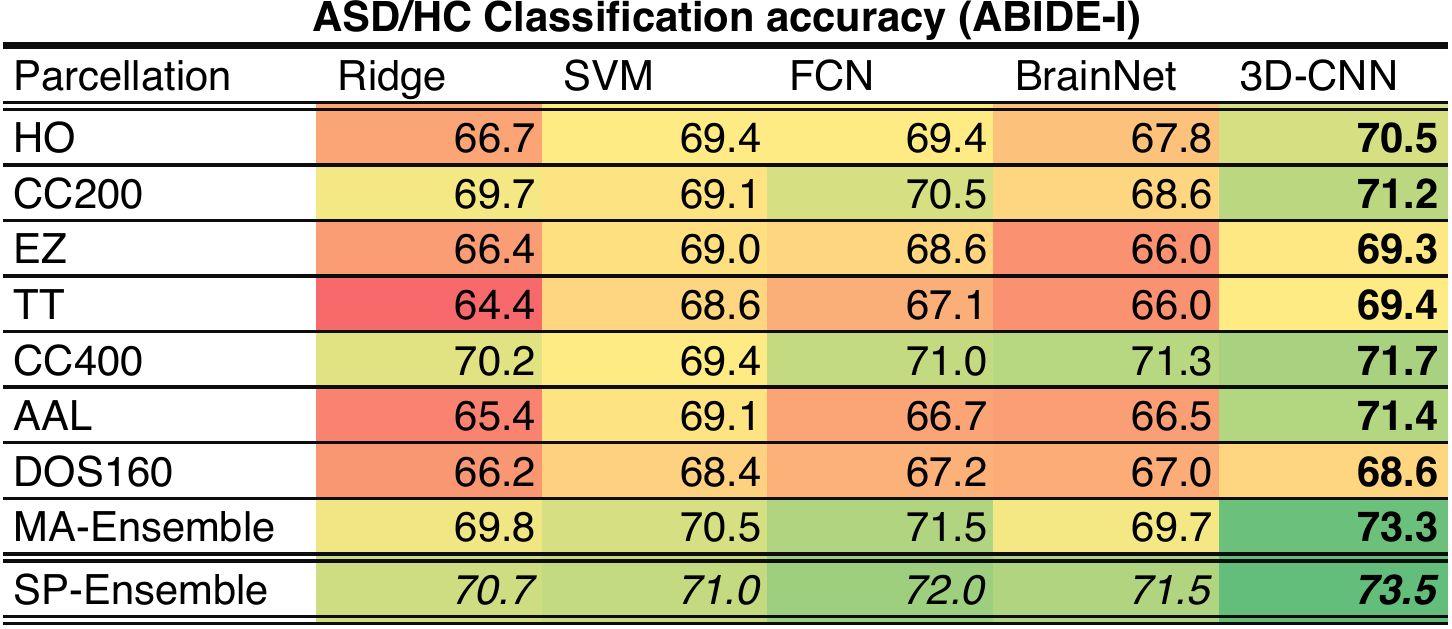}
\caption{Classification accuracy for ASD vs. Control: 10-fold cross-validation on ABIDE-I for benchmark models and proposed CNN approach. For each row, best results are \textbf{bolded}. For each column, best results are \textit{italicized}. Green indicates better performance, whereas orange/red highlights worse performance. 
}
\label{tab:abide1dx}
\end{table}

\begin{table}[t]
\centering\includegraphics[width=0.8\linewidth]{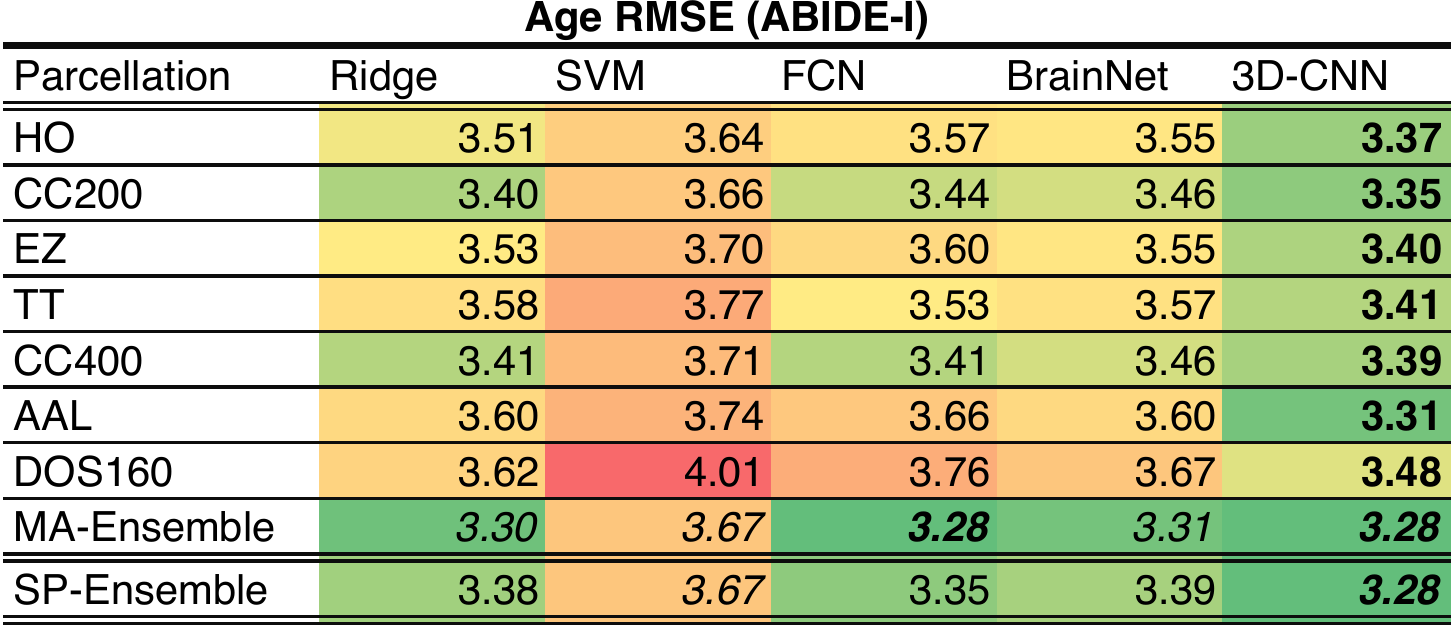}
\caption{Root mean squared error (RMSE in years) for age prediction: 10-fold cross-validation on ABIDE-I for benchmark models and proposed CNN approach. For each row, best results are \textbf{bolded}. For each column, best results are \textit{italicized}. Green indicates better performance, whereas orange/red highlights worse performance. 
}
\label{tab:abide1age}
\end{table}

\begin{figure}[!htb]
%\hspace*{-0.4in}
\begin{minipage}{0.9\textwidth}  
\includegraphics[width=1.2\linewidth]{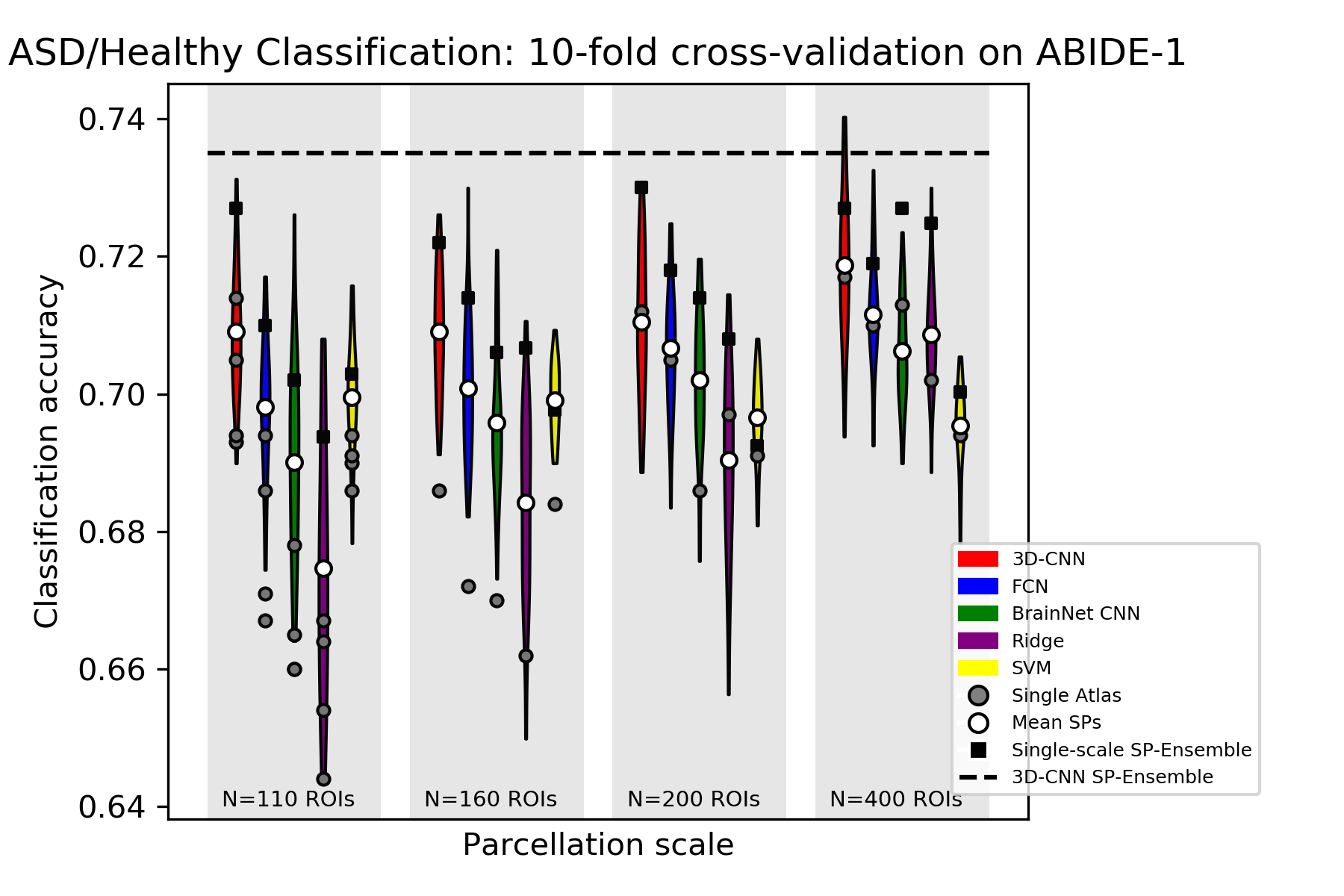}
%\caption{}
%\label{fig:cnn_violins_abide1_age}
\end{minipage}
%\hspace*{-0.4in}
\begin{minipage}{0.9\textwidth}  
\includegraphics[width=1.2\linewidth]{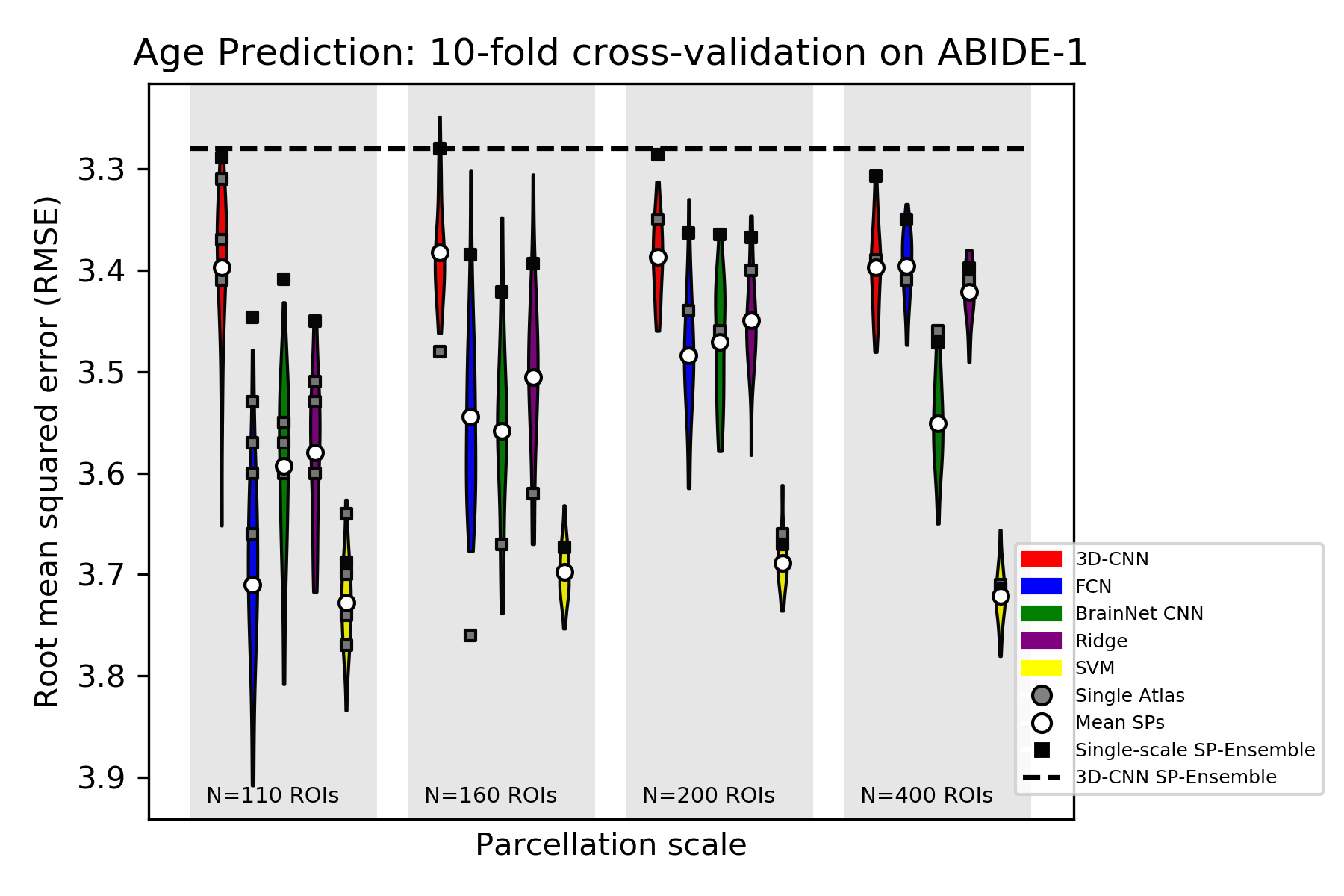}
%\caption{}
%\label{fig:cnn_violins_abide1_age}
\end{minipage}
\caption{Violin plots showing the spread of prediction accuracies/errors for stochastic parcellations at multiple network scales for different classification models. Mean accuracy/error of individual violins is denoted by 'Mean SPs'. Performance of individual atlases is compared with SPs with the closest \# of ROIs and is denoted as 'Single Atlas'. Results are computed by 10-fold cross-validation on the entire ABIDE-1 cohort.}
\label{fig:violins2}
\end{figure}

In order to ensure a fair comparison with other studies that report 10-fold cross-validation performance on ABIDE-I, we report the performance obtained using our benchmark and proposed models (along with the ensemble learning strategy) for both stochastic parcellations and atlases in the form of kernel density plots (Figure~ \ref{fig:violins2} and Tables~\ref{tab:abide1dx}, \ref{tab:abide1age}). Clearly, the results and conclusions on ABIDE-I remain consistent with ABIDE-II, with the 3D-CNN ensemble strategy outperforming all the baseline methods.   

%\clearpage 
%\subsection{ROC Curves for inidividual atlases}

\subsection{Saliency maps for individual parcellations}
\begin{figure}%[H]
  \hspace*{-0.5in}
  \includegraphics[width=1.1\textwidth]{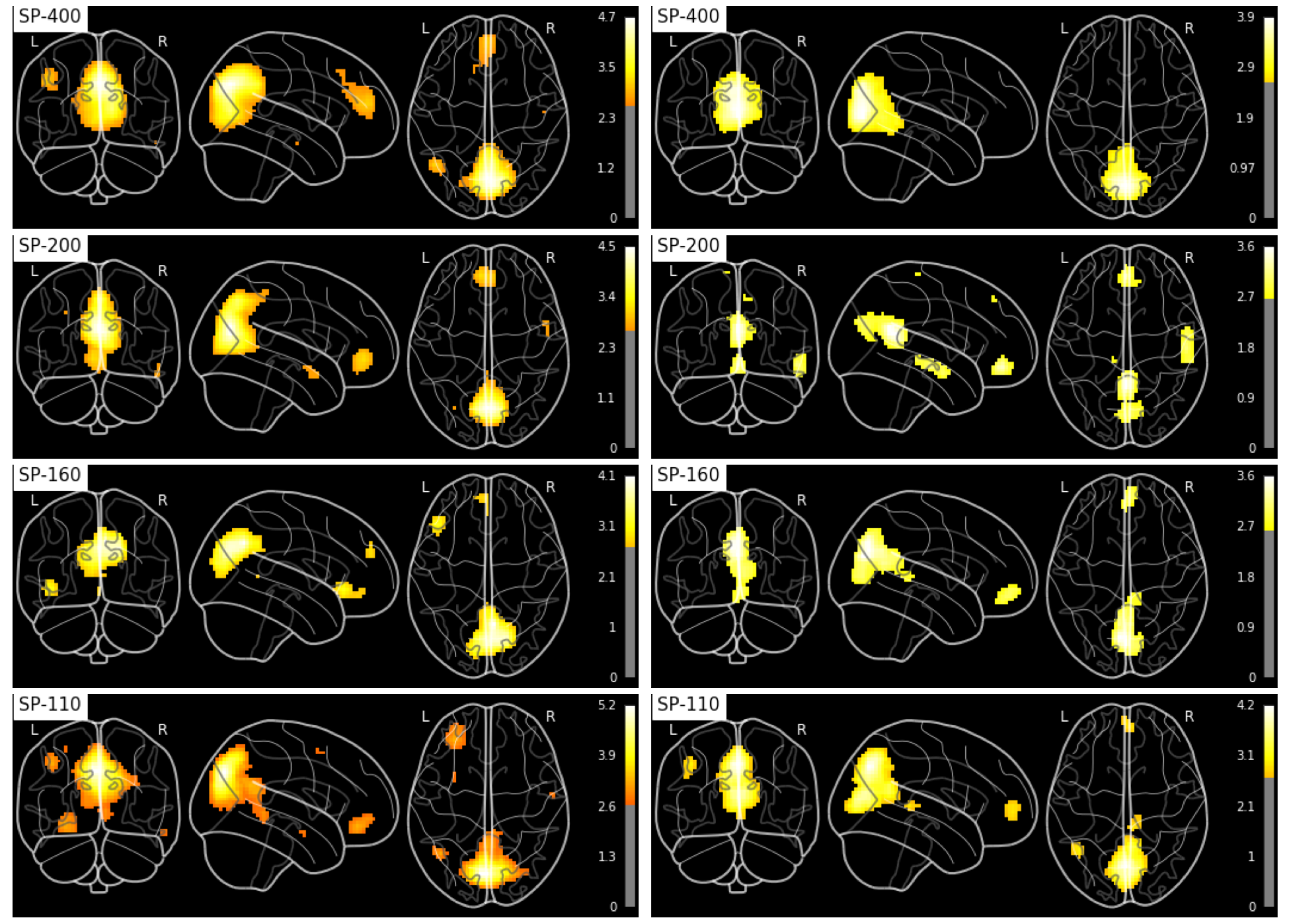} 
   \caption{Saliency maps of trained CNN models for 2 randomly chosen stochastic parcellations at each scale for ASD-HC classification.}
   \label{fig:saliency_sp_individual}
 \end{figure}

\begin{figure}%[H]
\hspace*{-0.5in}
\centering
\begin{subfigure}{.42\linewidth}
    \centering
     \includegraphics[width=3.5in]{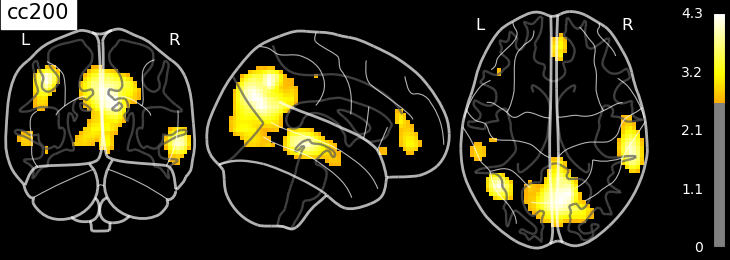} 
    %\caption{Image}\label{fig:cc200}
\end{subfigure}
    \hfill
\begin{subfigure}{.42\linewidth}
    \centering
    \includegraphics[width=3.5in]{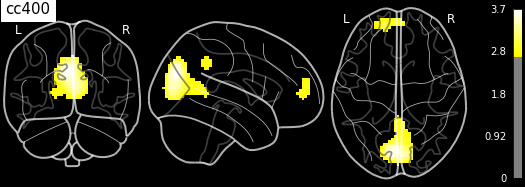} 
    %\caption{Image}\label{fig:cc400}
\end{subfigure}
   %\hfill

%\bigskip
\hspace*{-0.5in}
\centering 
\begin{subfigure}{.42\linewidth}
  \centering
  \includegraphics[width=3.5in]{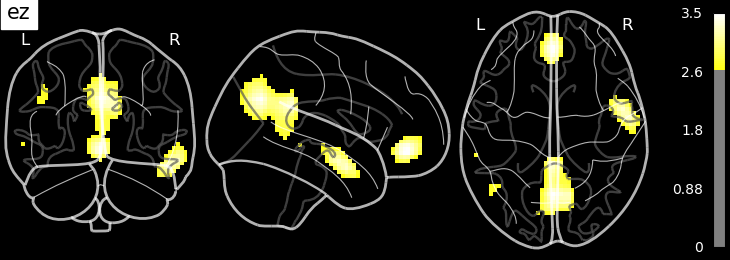} 
  %\caption{Image}\label{fig:ez}
\end{subfigure} 
\hfill
\begin{subfigure}{.42\linewidth}
    \centering
   \includegraphics[width=3.5in]{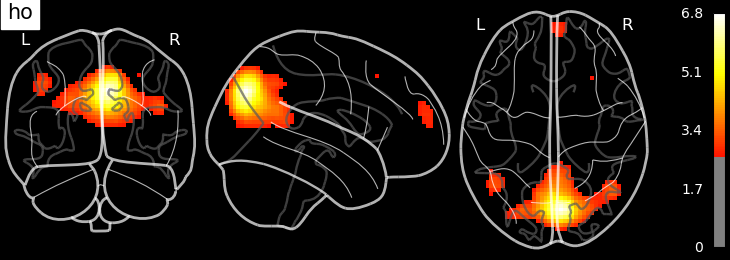} 
    %\caption{Image}\label{fig:ho}
\end{subfigure}
\hfill

%\bigskip
\hspace*{-0.5in}
\centering 
\begin{subfigure}{.42\linewidth}
  \centering
  \includegraphics[width=3.5in]{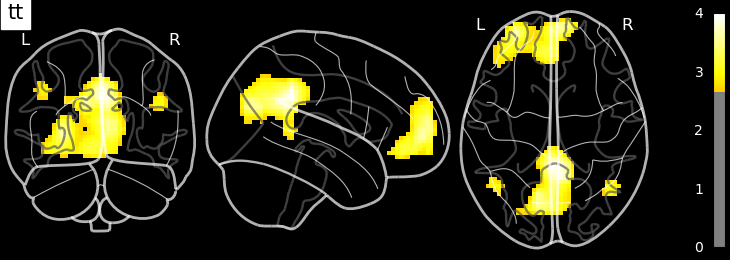} 
  %\caption{Image}\label{fig:tt}
\end{subfigure} 
\hfill
\begin{subfigure}{.42\linewidth}
    \centering
   \includegraphics[width=3.5in]{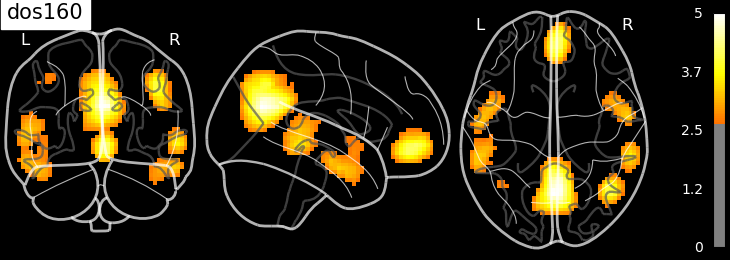} 
    %\caption{Image}\label{fig:dos160}
\end{subfigure}

%\bigskip
%\hspace*{-0.5in}
\centering 
\begin{subfigure}{\linewidth}
  \centering
  \includegraphics[width=3.5in]{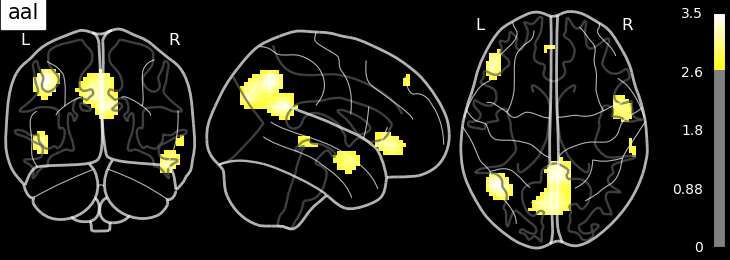} 
  %\caption{Image}\label{fig:aal}
\end{subfigure} 
\hfill

\caption{Saliency maps for atlas-based ASD-HC classification models.}
\label{fig:saliency_atlas}
\end{figure}

Visualizing the saliency maps for models trained on different brain parcellations can reveal interesting differences in the features captured by these models. We visualized the saliency maps of the 3D-CNN model for individual stochastic parcellations at multiple scales for the task of ASD/HC Classification. As shown in Figure~\ref{fig:saliency_sp_individual}, models trained using distinct parcellation schemes are relying on the same basic underlying connectivity patterns for prediction, with small differences in their information content, that can be utilized efficiently by the ensemble learning scheme. Further, the saliency maps of atlas-based (see Figure~\ref{fig:saliency_atlas}) and stochastic parcellation-based models are remarkably similar, suggesting that the connectivity patterns of the same set of voxels are guiding the classifier predictions, irrespective of the precise scheme of ROI extraction.

\subsection{Comparison of different preprocessing strategies}
Since preprocessing options such as nuisance regression have been a point of contention in several studies, we conducted another set of experiments with standard atlas masks in three preprocessing scenarios (a) without global signal regression(GSR) + with CompCor (b) without GSR + without CompCor and (c) with GSR + without CompCor. Below, we include the results obtained with models trained on the ABIDE-1 data using the hyperparameters optimized in our original experiments presented in the paper. The accuracy values were computed based on test predictions on the independent ABIDE-2 dataset, following our original evaluation protocol.   
As can be seen from Tables~ \ref{tab:preprocdx} and \ref{tab:preprocage}, when neither GSR nor CompCor is employed during preprocessing, the prediction performance on both the tasks, i.e., ASD/HC Classification and age prediction, drops significantly. However, similar performance is obtained when using either or both of CompCor and GSR in preprocessing. Importantly, the trend remains the same: the 3D-CNN fares favorably against the baseline algorithms and the MA-Ensemble models generally perform better than or similar to the best-performing atlas. 
%\subsubsection{Task 1: ASD/HC Classification}
\begin{table}[t]
\centering\includegraphics[width=0.8\linewidth]{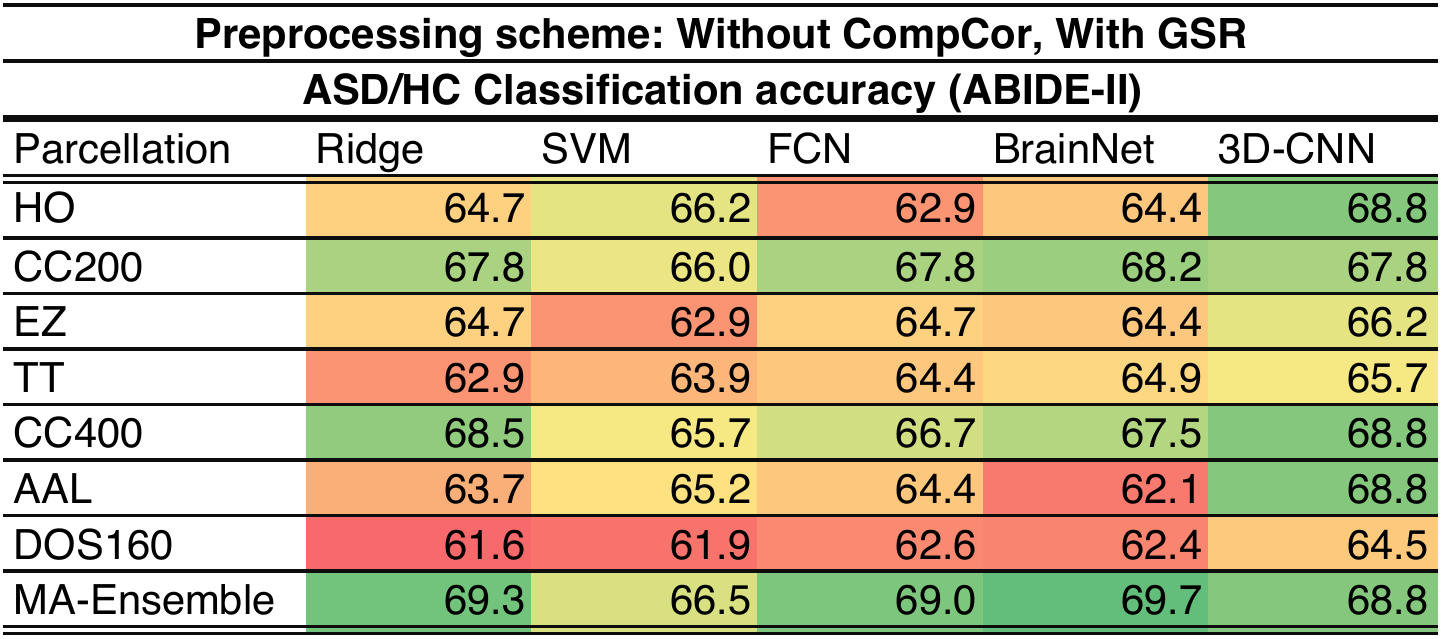}
\centering\includegraphics[width=0.8\linewidth]{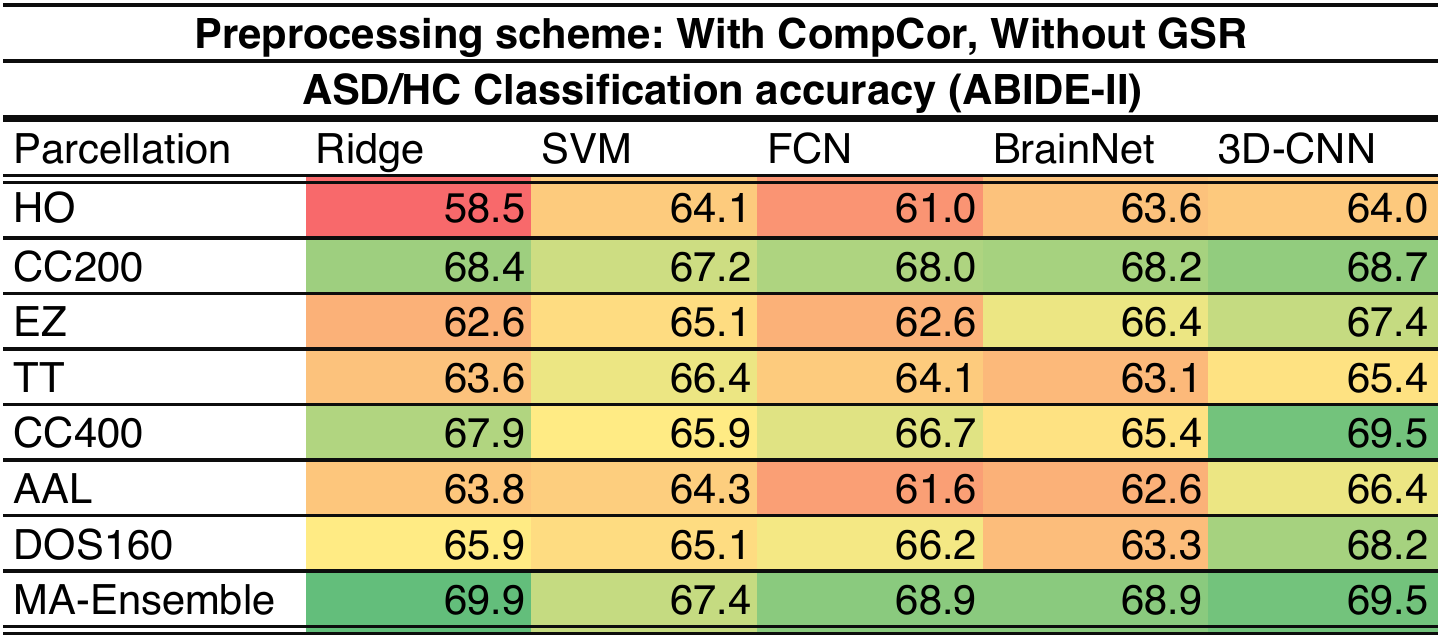}
\centering\includegraphics[width=0.8\linewidth]{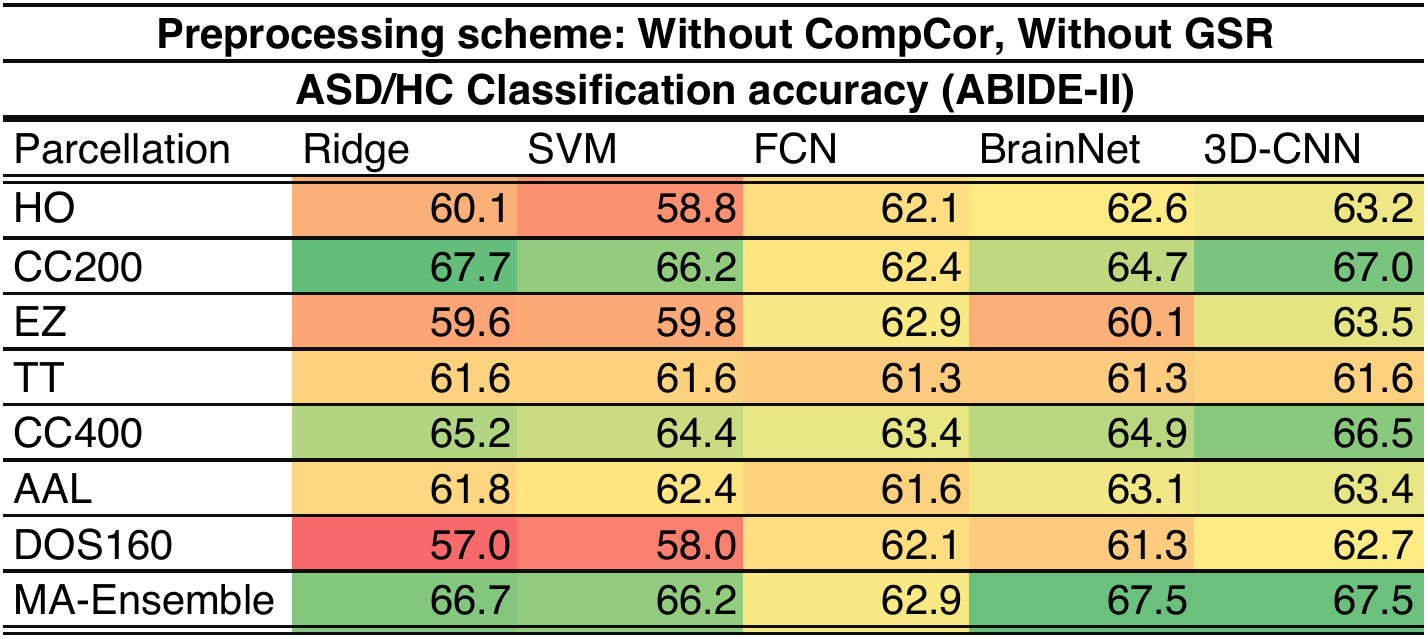}
\caption{Classification accuracy for ASD vs. Control: Independent results on ABIDE-II for benchmark models and proposed CNN approach. Green indicates better performance, whereas orange/red highlights worse performance. 
}
\label{tab:preprocdx}
\end{table}

%\subsubsection{Task 2: Age prediction}
\begin{table}[t]
\centering\includegraphics[width=0.8\linewidth]{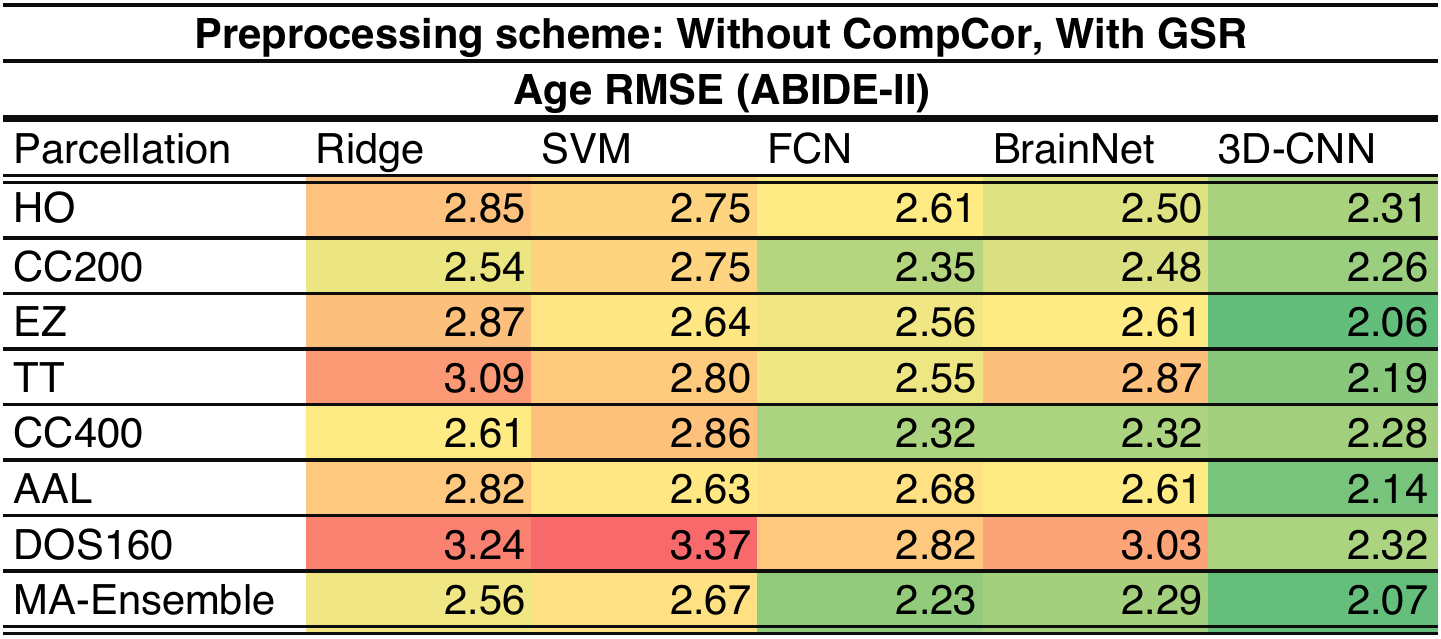}
\centering\includegraphics[width=0.8\linewidth]{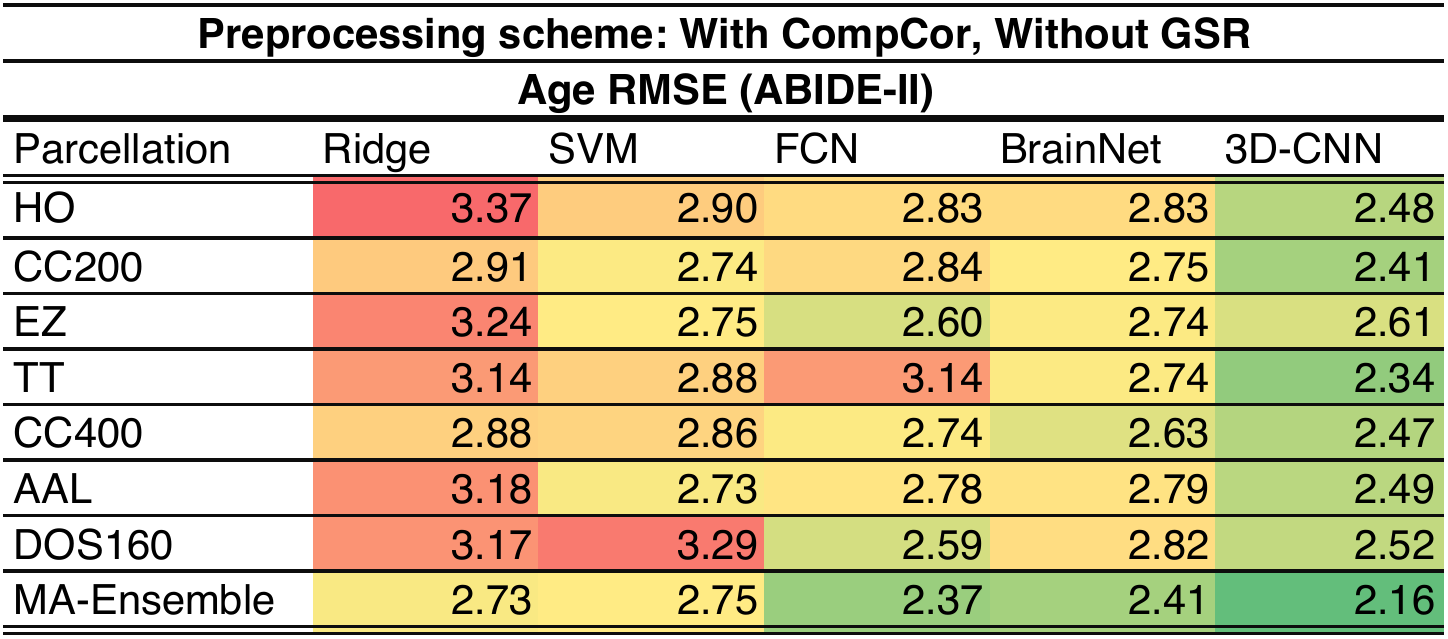}
\centering\includegraphics[width=0.8\linewidth]{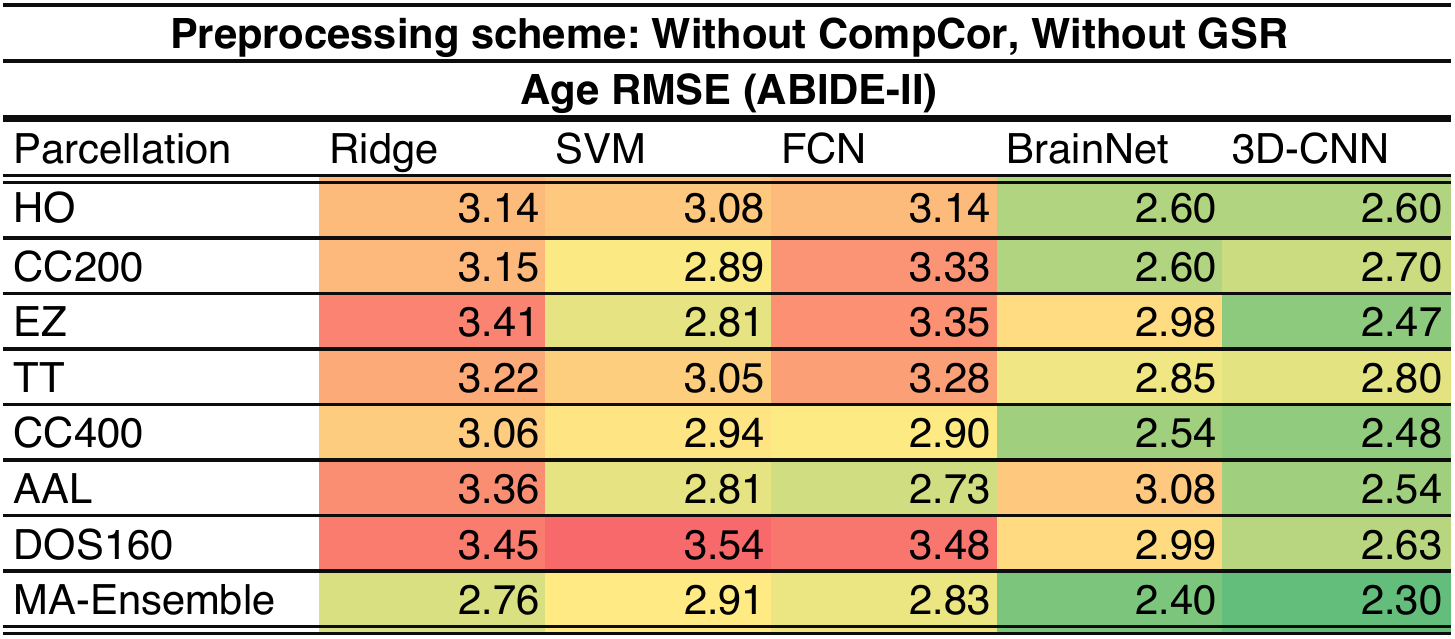}
\caption{Root mean squared error (RMSE in years) for age prediction: Independent results on ABIDE-II for benchmark models and proposed CNN approach. Green indicates better performance, whereas orange/red highlights worse performance. 
}
\label{tab:preprocage}
\end{table}

\begin{figure}%[H]
\hspace*{-0.5in}
\centering
\begin{subfigure}{.42\linewidth}
    \centering
     \includegraphics[width=3.2in]{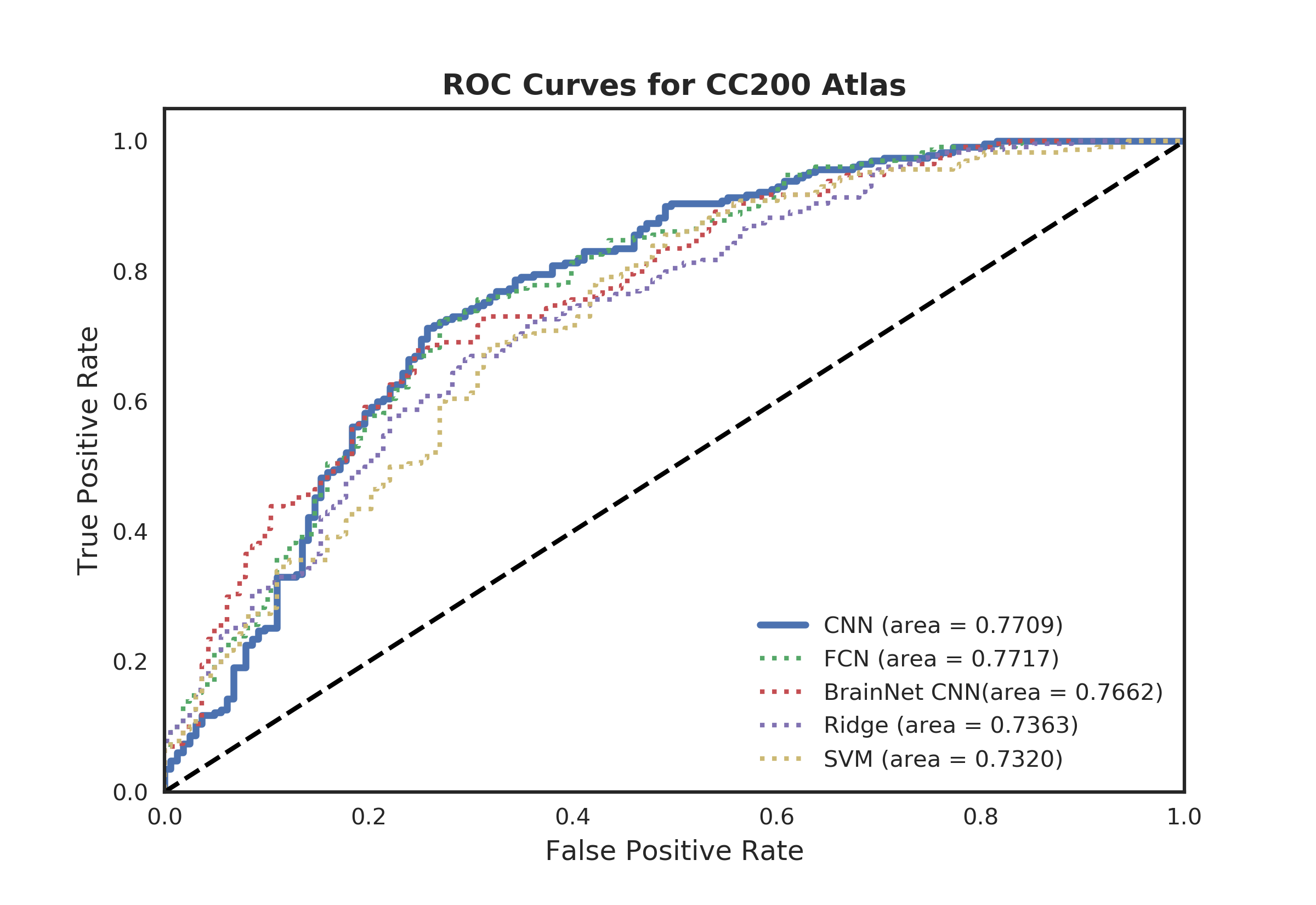} 
    %\caption{Image}\label{fig:cc200}
\end{subfigure}
    %\hfill
    \hspace*{0.8in}
\begin{subfigure}{.42\linewidth}
    \centering
    \includegraphics[width=3.2in]{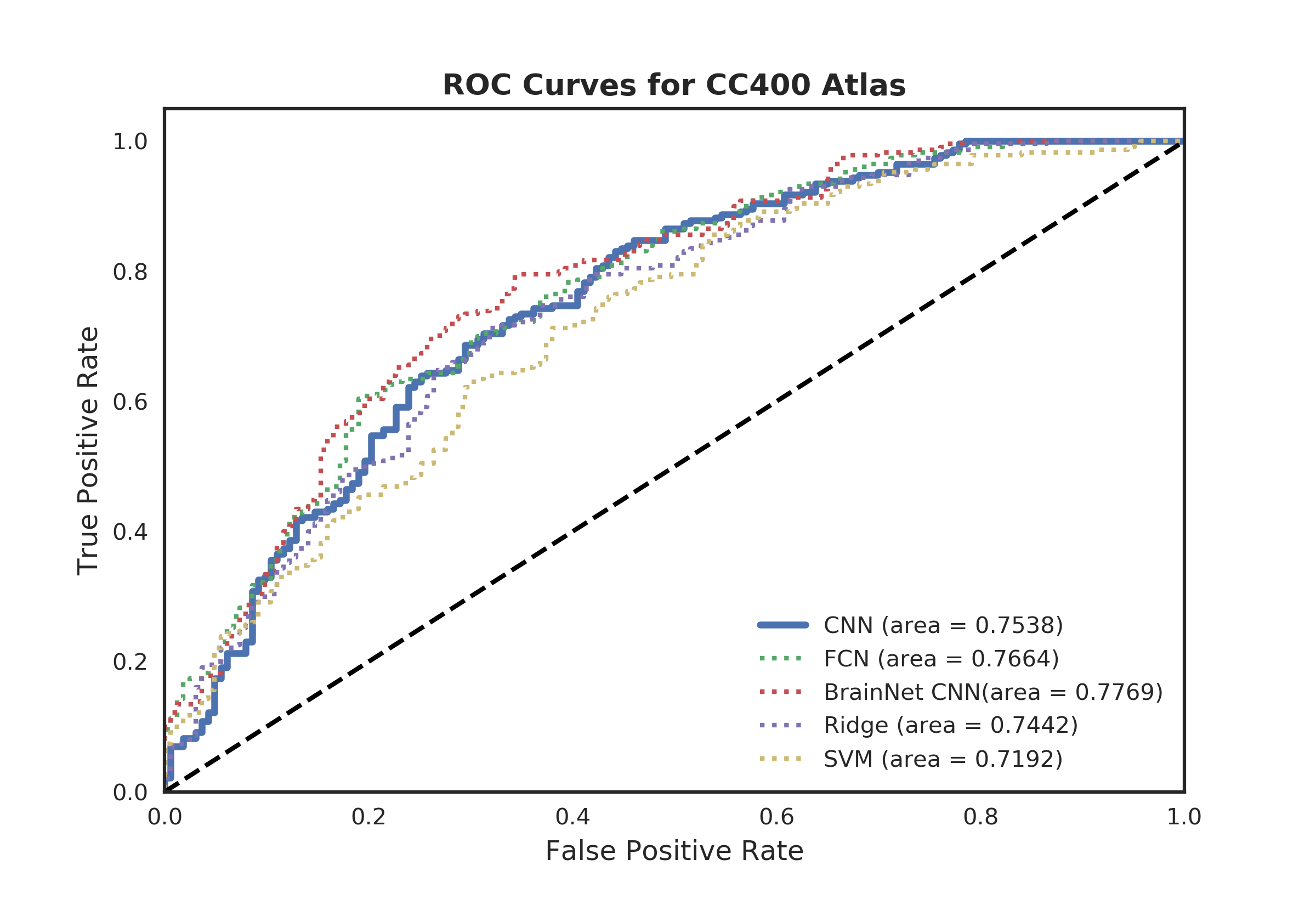} 
    %\caption{Image}\label{fig:cc400}
\end{subfigure}
%\hspace*{-0.3in}
   %\hfill

%\bigskip
\hspace*{-0.5in}
\centering 
\begin{subfigure}{.42\linewidth}
  \centering
  \includegraphics[width=3.2in]{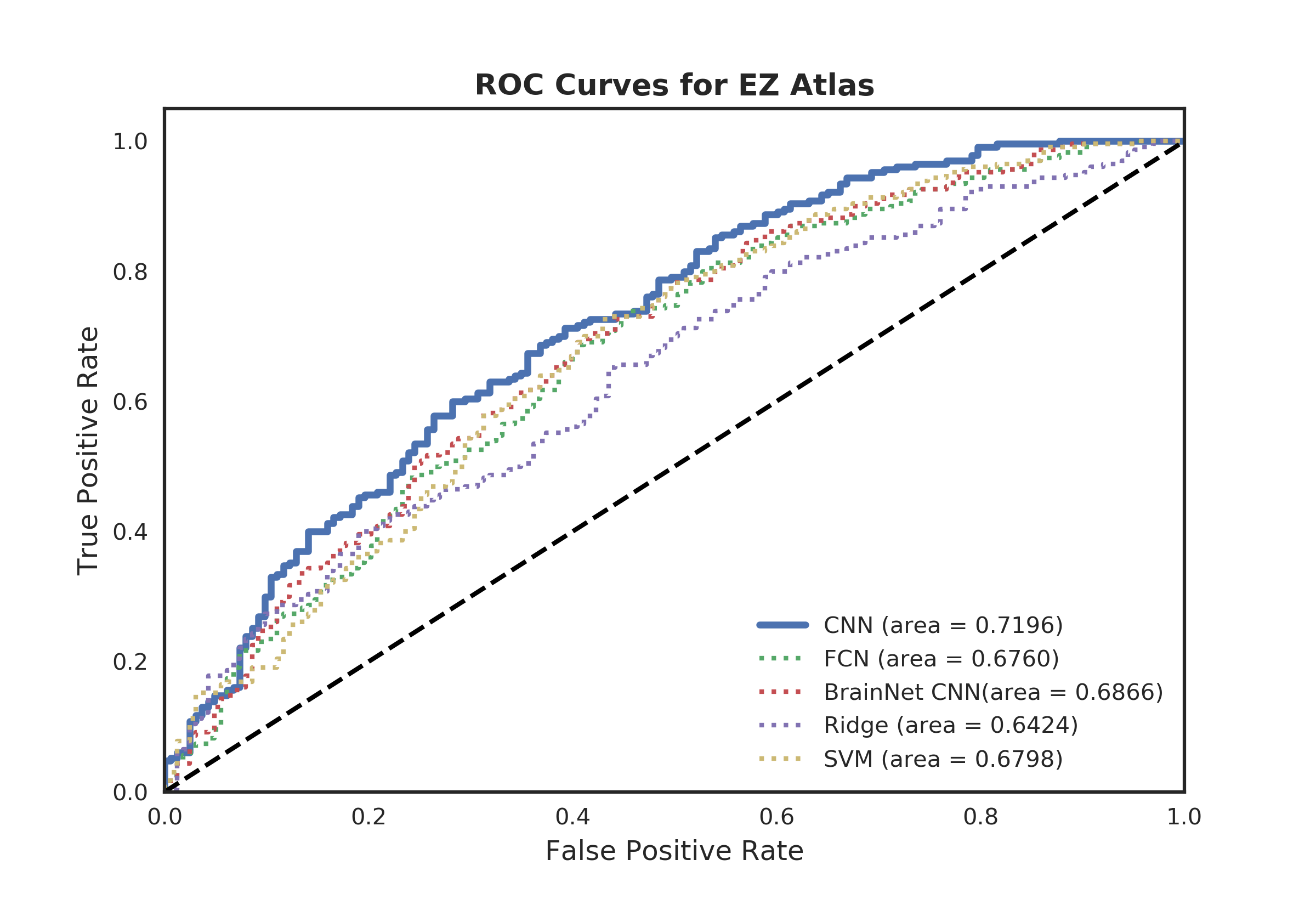}%\hspace{3em} 
  %\caption{Image}\label{fig:ez}
\end{subfigure} 
\hspace*{0.8in}
%\hfill
\begin{subfigure}{.42\linewidth}
    \centering
   \includegraphics[width=3.2in]{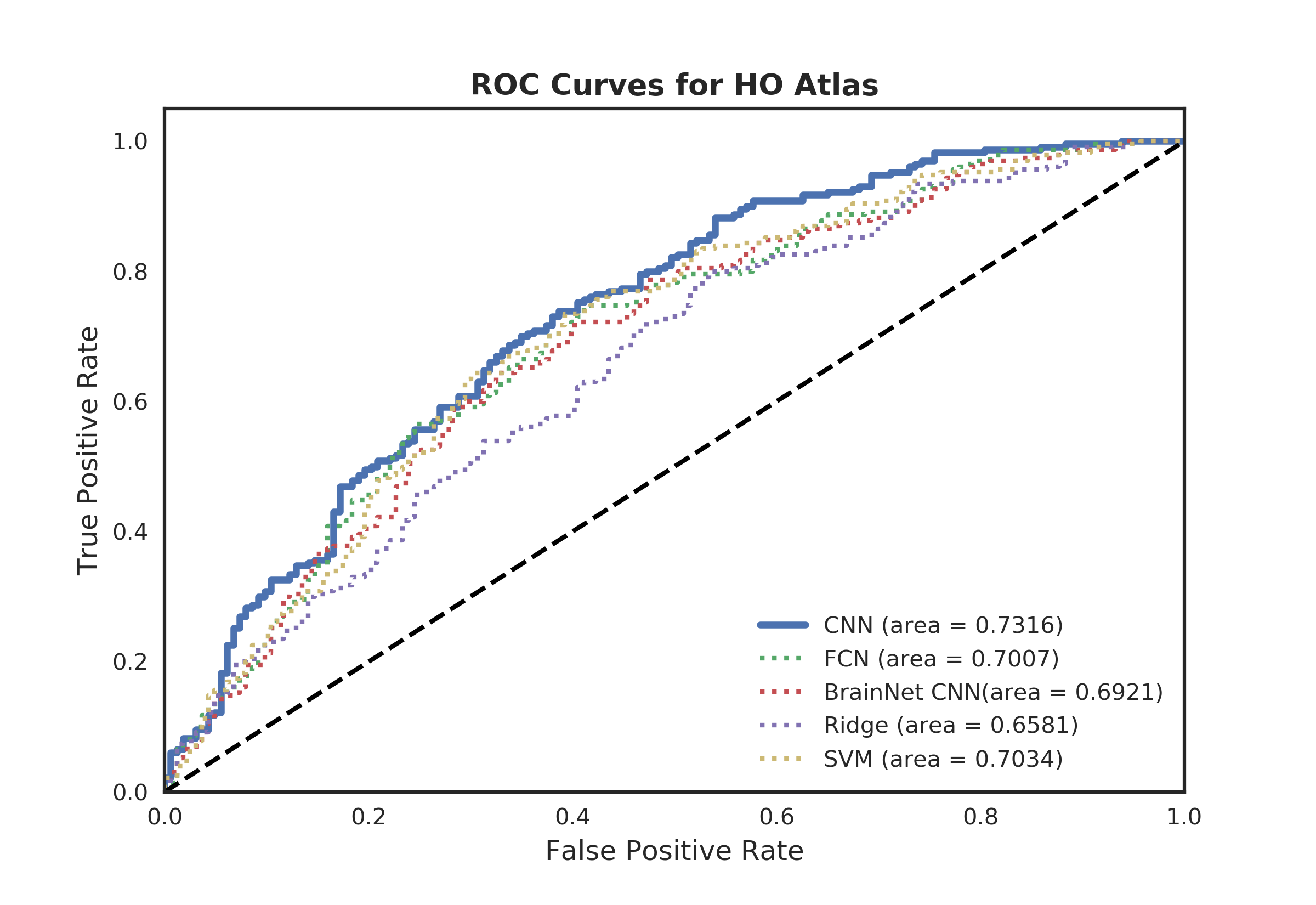} 
    %\caption{Image}\label{fig:ho}
\end{subfigure}
%\hspace*{-0.3in}
%\hfill
%\bigskip
\hspace*{-0.5in}
\centering 
\begin{subfigure}{.42\linewidth}
  \centering
  \includegraphics[width=3.2in]{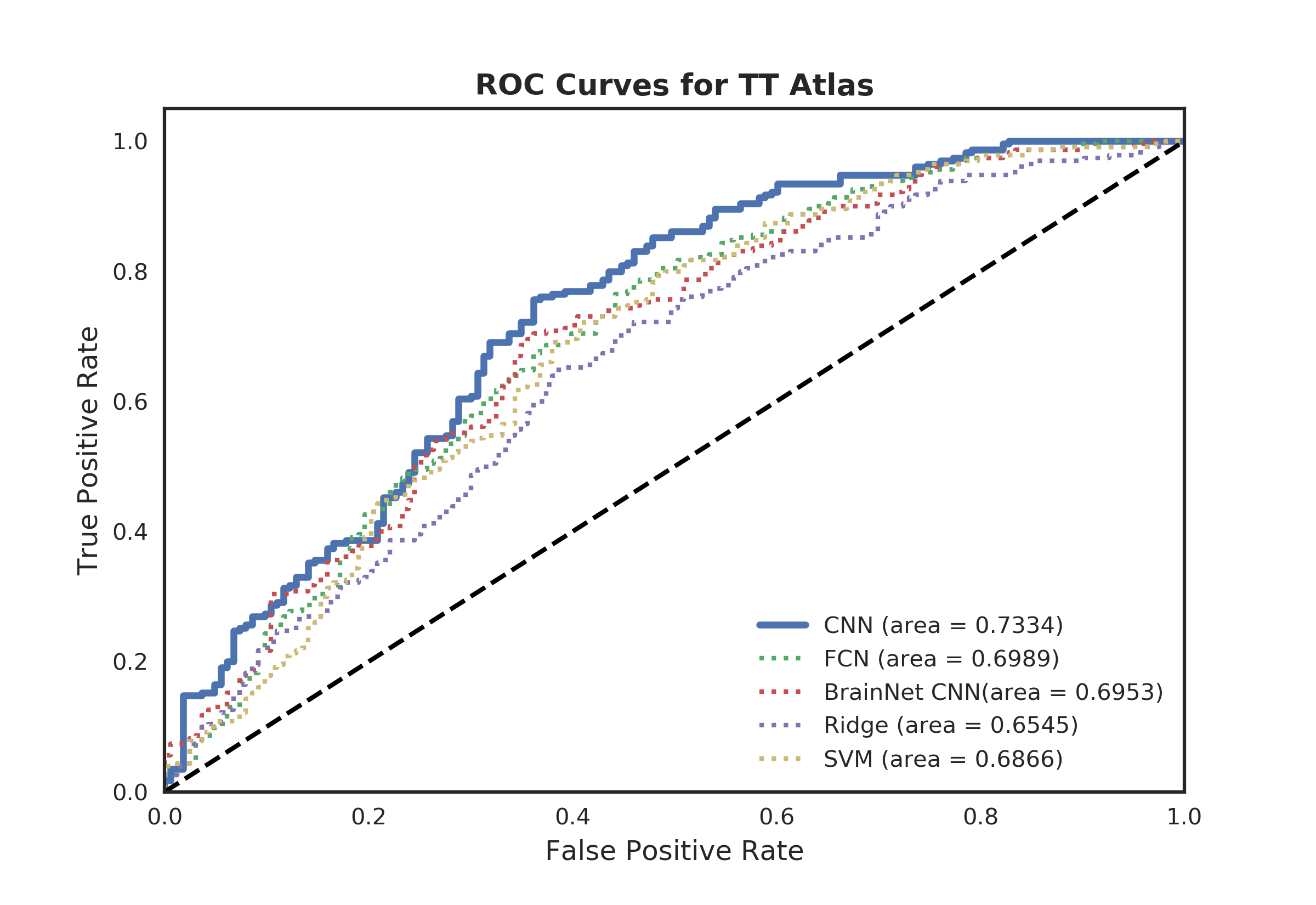} 
  %\caption{Image}\label{fig:tt}
\end{subfigure} 
\hspace*{0.8in}
%\hfill
\begin{subfigure}{.42\linewidth}
    \centering
   \includegraphics[width=3.2in]{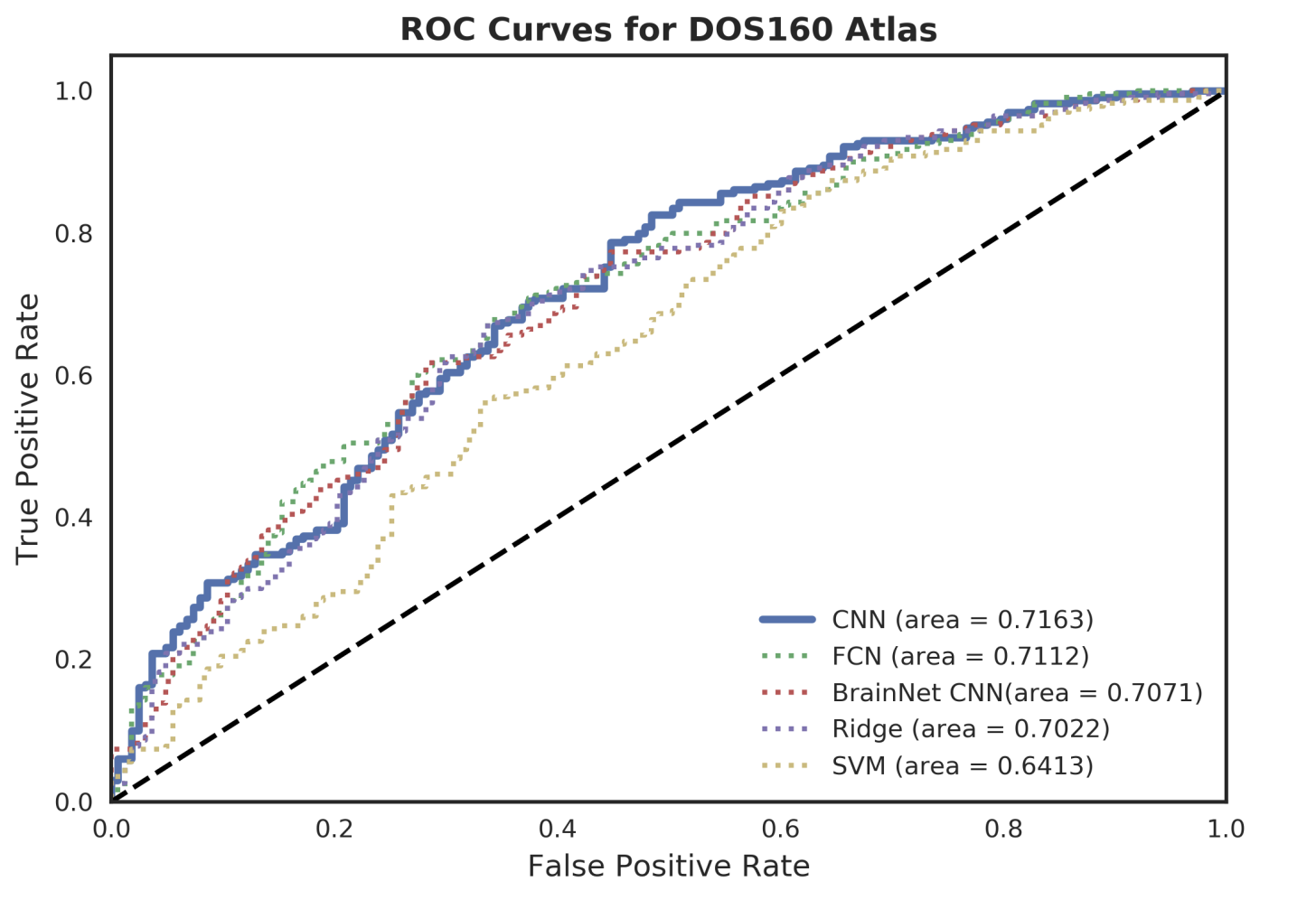} %\caption{Image}\label{fig:dos160}
\end{subfigure}
\hspace*{-0.3in}
%\bigskip
%\hspace*{-0.5in}
\hspace*{-0.5in}
\centering 
\begin{subfigure}{\linewidth}
  \centering
  \includegraphics[width=3.2in]{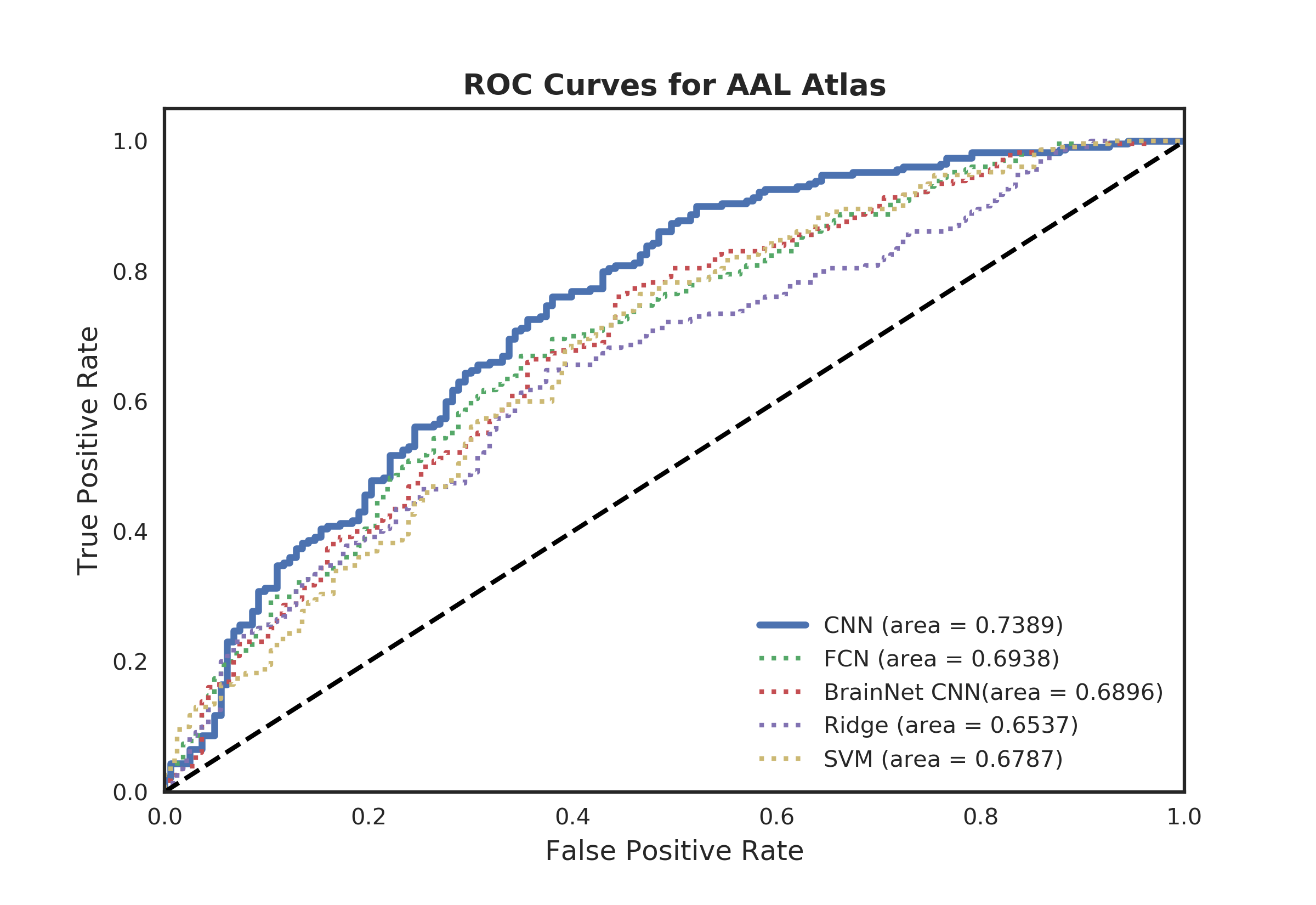} 
  %\caption{Image}\label{fig:aal}
\end{subfigure} 
\hspace*{-0.3in}
%\bigskip
%\hspace*{-0.5in}
\hspace*{-0.5in}
\centering 
\caption{ROC Curves for individual atlas based ASD-HC classification models.}
\label{fig:roc_atlases}

\end{figure}

%\subsection{Mean absolute error}

\begin{table}%[H]
\centering\includegraphics[width=0.8\linewidth]{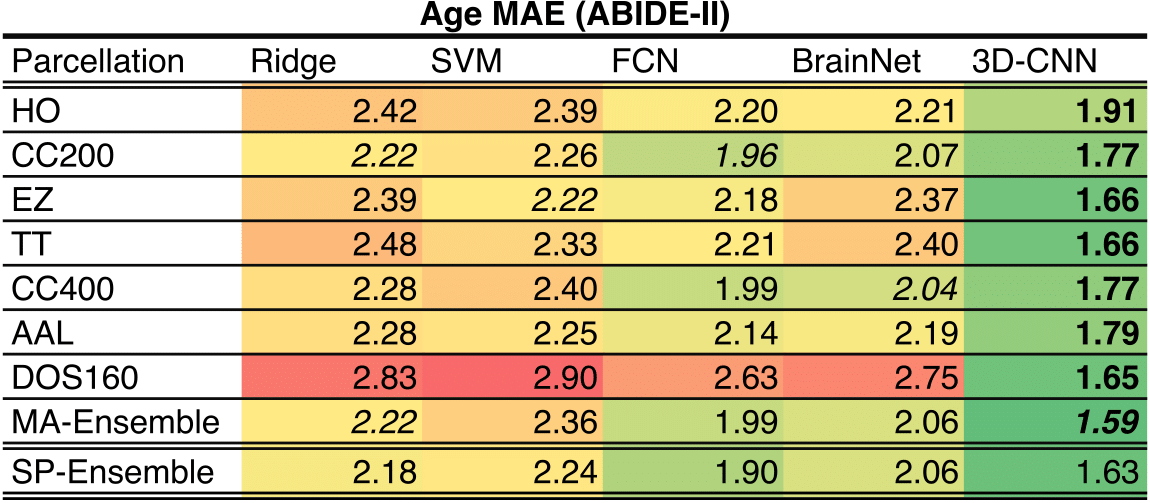}
\caption{Mean absolute error (MAE in years) for age prediction: Independent testing on ABIDE-II for benchmark models and proposed CNN approach. For each row, best results are \textbf{bolded}. For each column, best results are \textit{italicized}. Green indicates better performance, whereas orange/red highlights worse performance. 
}
\label{tab:abide2maeage}
\end{table}
\end{document}